\documentclass{article}

\usepackage[preprint]{neurips_2026}

\usepackage[utf8]{inputenc}
\usepackage[T1]{fontenc}
\usepackage{amsmath,amssymb,amsfonts}
\usepackage{graphicx}
\usepackage{booktabs}
\usepackage{multirow}
\usepackage{colortbl}
\usepackage{hyperref}
\usepackage{cleveref}
\usepackage{xcolor}
\usepackage{tikz}
\usepackage{pgfplots}
\usepackage{enumitem}
\pgfplotsset{compat=1.18}
\usetikzlibrary{arrows.meta,positioning,shapes.geometric,calc,fit,decorations.pathreplacing}

\definecolor{cGray}{HTML}{BDBDBD}      
\definecolor{cBlueLight}{HTML}{9ECAE1}  
\definecolor{cBlueMed}{HTML}{4292C6}    
\definecolor{cBlueDark}{HTML}{2171B5}   
\definecolor{cBlueDeep}{HTML}{084594}   
\definecolor{cOrange}{HTML}{E69F00}     
\definecolor{cPosL}{HTML}{EDF7EC}       
\definecolor{cPos}{HTML}{D5E8D4}        
\definecolor{cPosM}{HTML}{A9D4A7}       
\definecolor{cPosH}{HTML}{7DC07D}       
\definecolor{cNeg}{HTML}{F8D7DA}        
\colorlet{cStateless}{cGray}
\colorlet{cTable}{cBlueLight}
\colorlet{cEpisodic}{cBlueMed}
\colorlet{cEpiMiss}{cBlueDark}
\colorlet{cFull}{cBlueDeep}
\newcommand{\posL}[1]{\cellcolor{cPosL}#1}
\newcommand{\pos}[1]{\cellcolor{cPos}#1}
\newcommand{\posM}[1]{\cellcolor{cPosM}#1}
\newcommand{\posH}[1]{\cellcolor{cPosH}#1}
\newcommand{\nega}[1]{\cellcolor{cNeg}#1}

\title{Speculate with Memory: Lossless Acceleration for LLM Agents}

\author{
  Yu Li \\
  Salesforce Research \\
  \texttt{yu.li@salesforce.com}
  \And
  Qinyuan Ye \\
  Salesforce Research \\
  \texttt{qinyuan.ye@salesforce.com}
  \And
  Prafulla Kumar Choubey \\
  Salesforce Research \\
  \texttt{pchoubey@salesforce.com}
  \AND
  Jiaxin Zhang \\
  Salesforce Research \\
  \texttt{jiaxin.zhang@salesforce.com}
  \And
  Chien-Sheng Wu \\
  Salesforce Research \\
  \texttt{wu.jason@salesforce.com}
}

\begin{document}

\maketitle


\begin{abstract}
Speculative execution accelerates LLM agents by using a smaller, cheaper model to predict and pre-launch the next step while the environment is idle. However, existing speculators are stateless and discard all information between tasks, preventing prediction quality from improving with experience. We equip the speculator with three online memory systems that learn from past agent trajectories: a contrastive transition table tracking action-sequence statistics, an episodic memory retrieving contextually similar segments, and a confusion tracker suppressing recurring errors. We evaluate this approach on six benchmarks spanning three speculation types: action prediction, observation prediction, and chained prediction. Memory-augmented speculation yields a 19--39\% relative accuracy improvement on action prediction and up to a $2.5\times$ increase on observation prediction tasks with repetitive action spaces. These gains grow continuously as memory accumulates and generalize across speculator models of varying cost. All speculation is lossless because it runs during idle time at zero added wall-clock cost, and the actor's trajectory is identical to non-speculative execution.
\end{abstract}


\section{Introduction}
\label{sec:intro}

Large language model (LLM) agents interact with external environments through tool calls, browser actions, and physical commands, where each step incurs latency from both the LLM inference and the action execution.  In interactive settings such as customer service, web navigation, and embodied control, this sequential latency accumulates and directly affects user experience. Speculative execution addresses this bottleneck by using a smaller, faster model to predict the agent's next action and pre-launch it before the agent commits~\citep{ye2025speculative}. The idea draws on speculative execution in processor architecture and speculative decoding in LLM inference~\citep{leviathan2023speculative}. If the prediction is correct, the result is already available and latency is hidden. If incorrect, the pre-launched work is discarded at no functional cost. However, most speculative execution methods for LLM agents are \textbf{stateless}, treating every task independently. Recent work mines tool-calling patterns offline~\citep{sui2026paste}, but no existing method retrieves contextually similar past trajectories or adapts to its own prediction failures during deployment. A customer-service speculator that has processed 500 prior interactions makes the same predictions as one that has seen none. Yet agent tasks exhibit strong regularity. The same action sequences recur across episodes: looking up a user followed by retrieving their order in customer service, or navigating to a shelf followed by picking up a mug in household tasks. A system that remembers these patterns should predict more accurately over time.

We augment the speculator with structured memory that accumulates online from past agent trajectories. Drawing on recent advances in agent memory systems~\citep{wang2024awm,zhao2024expel,ouyang2025reasoningbank}, we equip the speculator with three complementary components that capture aggregate workflow patterns, retrieve contextually similar past episodes as in-context examples, and correct recurring prediction errors. While the actor is busy, the speculator predicts the outcome and pre-launches the next step along each predicted branch. When the actor's real output arrives, the system checks whether any speculative branch matches. On a hit, the pre-launched work is already complete, so the agent advances immediately instead of issuing a fresh blocking call. On a miss, the speculative work is discarded and execution proceeds normally, so the actor's trajectory is identical to non-speculative execution. Pre-launched work is restricted to side-effect-free operations (LLM calls, read-only API queries, page fetches), ensuring that incorrect predictions never produce irreversible changes and preserving the lossless guarantee (\Cref{sec:method:framework}). Only pre-launchable predictions, those whose execution can overlap with and hide latency, translate into wall-clock savings. We evaluate seven progressive memory settings on six benchmarks spanning web navigation~\citep{zhou2024webarena,koh2024vwa}, embodied control~\citep{shridhar2021alfworld}, planning~\citep{vallati2015ipc}, customer service~\citep{yao2024tau, barres2025tau2}, and multi-hop QA~\citep{yang2018hotpotqa}.

\input{figures/intro_figure}
Memory-augmented speculation consistently outperforms the stateless baseline across all six benchmarks. The magnitude varies by speculation type: 19--39\% relative improvement on action-prediction tasks and up to $2.5\times$ on observation-prediction tasks with structured action spaces. \Cref{fig:intro_figure}~(left) illustrates the mechanism. During environment idle time (API calls, user responses, page loads), the speculator predicts the next output and pre-launches it. Memory of past trajectories increases the hit rate. \Cref{fig:intro_figure}~(right) shows the estimated latency reduction on ALFWorld over 120 episodes. Each correct speculation hides the latency of one actor LLM call, and the memory-augmented speculator improves from $\sim$28\% to over 50\% as it accumulates experience, while the stateless baseline remains flat. The improvements are model-agnostic and generalize across speculator models of varying capability and cost, and the optimal memory configuration varies by domain. Our contributions are as follows:
\begin{itemize}[nosep,leftmargin=*]
    \item We augment speculative execution for LLM agents with three complementary memory systems that accumulate experience from past trajectories: a contrastive transition table, an episodic memory with miss-episode recording, and a confusion tracker.
    \item We identify three speculation types based on where latency falls in the agent-environment loop, unifying prior work under a single framework.
    \item We evaluate seven progressive memory settings across six benchmarks and three speculation types, showing consistent improvement over stateless speculation, with the largest gains on observation-prediction benchmarks with structured action spaces (up to $2.5\times$).
\end{itemize}


\section{Speculate with Memory}
\label{sec:method}

\subsection{Speculation Framework}
\label{sec:method:framework}
Consider an LLM agent interacting with an environment in a turn-based loop. At each turn~$t$, the agent observes state~$s_t$, comprising the conversation history and environment feedback, and selects an action~$a_t = (\textit{type}, \textit{args})$ via an LLM call with latency~$\ell_{\text{LLM}}$. The environment returns observation~$o_t$ with latency~$\ell_{\text{env}}$, which encompasses waiting for API responses, user input, or environment state computation. A \emph{speculator} is a smaller, faster model that predicts future outputs in this loop. Let $\ell_{\text{spec}}$ denote the speculator's inference latency. Depending on the prediction target, the speculator runs either during environment processing or in parallel with the actor. In both cases, its cost is fully hidden because it overlaps with ongoing computation. With $k$~parallel speculators, a best-of-$k$ strategy succeeds if any single prediction matches.

\begin{figure*}[t]
\centering
\begin{minipage}[t]{0.42\textwidth}
\centering
\includegraphics[width=\textwidth]{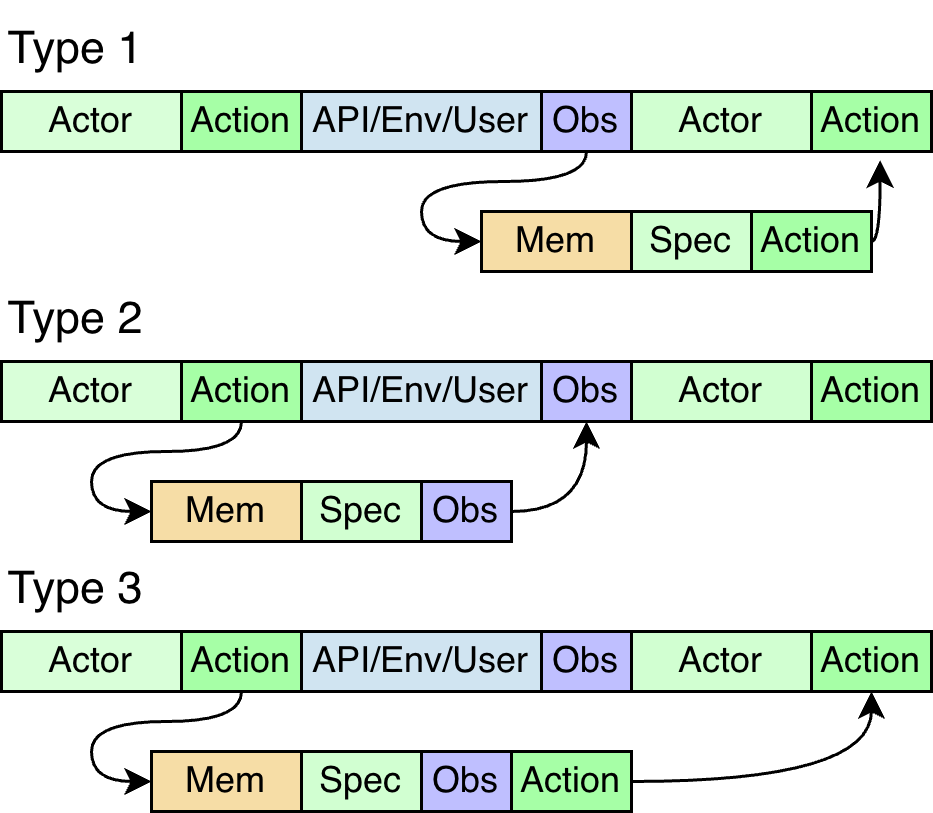}
\end{minipage}%
\hfill
\begin{minipage}[t]{0.56\textwidth}
\centering
\includegraphics[width=\textwidth]{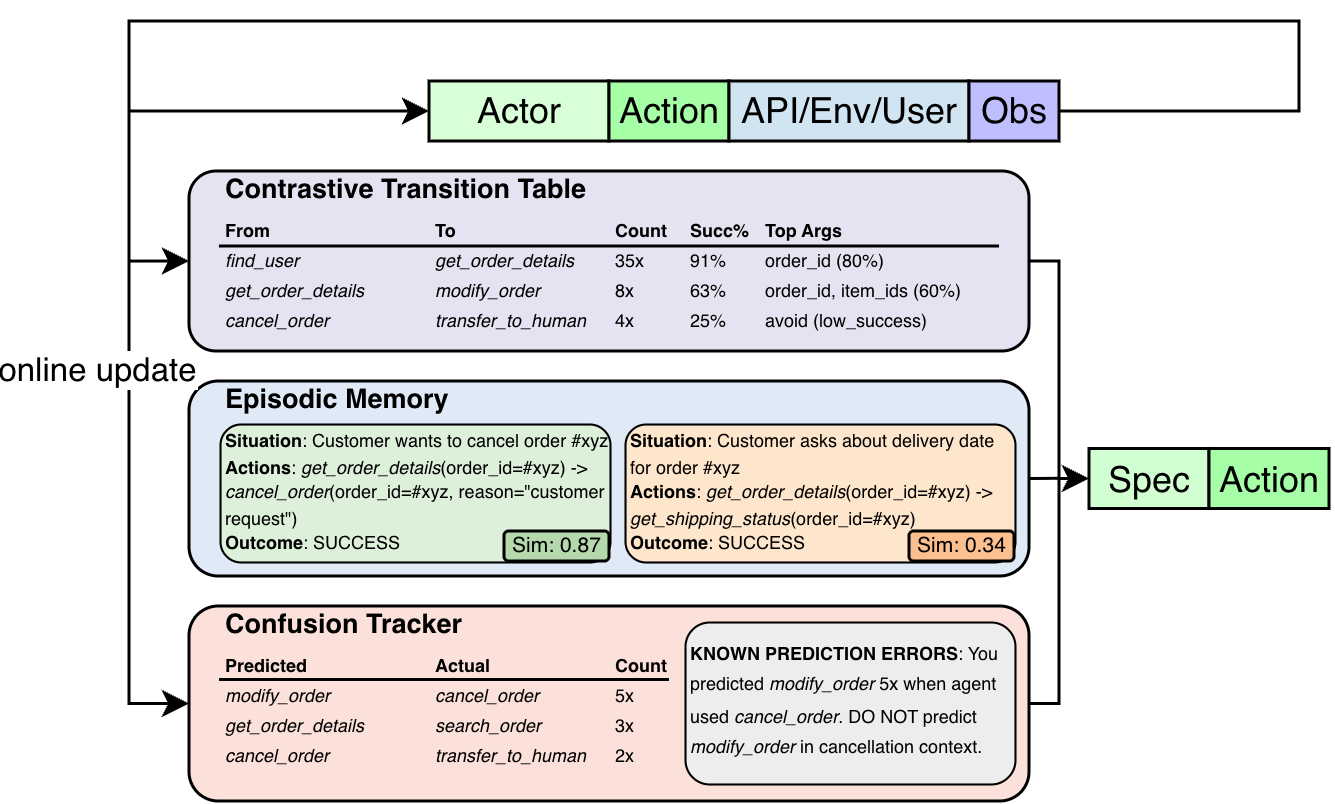}
\end{minipage}
\caption{\textbf{Left:}~Timing of the three speculation types.  Each row
shows the agent-environment loop (top) and the speculator's parallel work
(bottom); shaded regions indicate saved latency on a hit.
\textbf{Right:}~Memory-augmented speculation architecture.  The agent loop
feeds online updates to three memory components (contrastive transition
table, episodic memory, and confusion tracker) that provide structured
context to the speculator.  Examples are from $\tau^{2}$-bench.}
\label{fig:method_overview}
\end{figure*}

What to predict, and what latency is saved on a hit, depends on the structure of the agent-environment loop. We identify three speculation types. In \emph{action prediction} (Type~1), the speculator predicts the agent's next action~$\hat{a}_{t+1}$ in parallel with the actor. On a hit, the predicted action was already pre-launched, saving up to~$\ell_{\text{env}}$. The realized saving is $\min(\ell_{\text{env}},\; \ell_{\text{LLM}} - \ell_{\text{spec}})$ since the speculator must finish within the actor's $\ell_{\text{LLM}}$~window to pre-launch. In \emph{observation prediction} (Type~2), the speculator predicts the environment's next observation~$\hat{o}_t$ during environment processing. On a hit, the actor's next LLM call was already pre-started with the predicted observation, saving up to~$\ell_{\text{LLM}}$. The realized saving is $\min(\ell_{\text{LLM}},\; \ell_{\text{env}} - \ell_{\text{spec}})$. Therefore, Type~2 is most effective when environment feedback is slow relative to the speculator, such as in robot execution, user input delays, or heavy API calls. In \emph{chained prediction} (Type~3), the speculator predicts both the observation and the next action during environment processing, chaining Type~2 and Type~1 in a single round. On a full hit, both the actor call and the action are pre-computed, saving up to $\ell_{\text{LLM}} + \ell_{\text{env}}$. In all cases, the saving approaches its upper bound when $\ell_{\text{spec}} \ll \ell_{\text{window}}$, which motivates using a smaller, faster speculator model. \Cref{fig:method_overview}~(left) illustrates the three types. The choice of type is determined by which steps incur significant latency and whether the environment's response is tractable to predict.

A correct prediction is useful only if the predicted work can be safely initiated before verification completes. For Type~2, the pre-launched work is an actor LLM call with predicted input. This is always safe since no environment state is modified. For Types~1 and~3, the pre-launched work is action execution, which is safe only for read-only actions such as API queries, page navigations, and database reads. Write actions with irreversible side effects are excluded. In practice, each benchmark defines a static whitelist of read-only action types derived from its API schema or action space. The speculator's predicted action type is checked against this whitelist before pre-launching. In all cases, we require a full exact match. Both the action type and all parameters must agree, because wrong parameters cannot be safely pre-launched.

\subsection{Memory Systems}
\label{sec:method:memory}
We equip the speculator with three complementary memory systems that learn from past agent trajectories (\Cref{fig:method_overview}, right): a contrastive transition table, an episodic memory, and a confusion tracker.

\paragraph{Contrastive Transition Table.}
Agent trajectories contain reusable action patterns~\citep{wang2024awm,fang2025memp,wang2023voyager}. We capture these in a transition table~$\mathcal{T}$ indexed by action-type pairs $(a_i, a_j)$. For each observed transition in a completed trajectory $\tau = [(a^{(1)}, \text{args}^{(1)}), \ldots, (a^{(n)}, \text{args}^{(n)})]$ with outcome label~$y \in \{+, -\}$, we record:

\begin{equation}
\begin{split}
    \forall\; 1 \le t < n: \quad
    & n^{y}_{a^{(t)}, a^{(t+1)}} \mathrel{+}= 1,
    \text{ArgSig}(a^{(t)}, a^{(t+1)}) \mathrel{\leftarrow} \text{ArgSig}(a^{(t)}, a^{(t+1)}) \\
    & \qquad \cup \{\text{sorted\_keys}(\text{args}^{(t+1)})\}
\end{split}
\end{equation}

where $n^{+}_{ij}$ and $n^{-}_{ij}$ are success and failure counts, and $\text{ArgSig}$ accumulates the sorted parameter names observed for action~$a_j$ when preceded by~$a_i$ (e.g., \texttt{order\_id,reason}). The table provides transition confidence and per-transition success rate:

\begin{equation}
    \text{Confidence}(a_i \to a_j) = \frac{n^{+}_{ij} + n^{-}_{ij}}{\sum_k (n^{+}_{ik} + n^{-}_{ik})} \qquad
    \text{SuccessRate}(a_i \to a_j) = \frac{n^{+}_{ij}}{n^{+}_{ij} + n^{-}_{ij}}
\end{equation}

Storing success and failure counts separately provides a contrastive signal. This is motivated by work showing that contrasting successful versus failed trajectories yields richer learning signals~\citep{zhao2024expel,ouyang2025reasoningbank}, and enables the speculator to prioritize high-confidence, high-success-rate continuations. The table is fundamentally unconditional: it captures $P(\text{next\_action} \mid \text{current\_action})$ but not the dependence on conversational context or specific argument values.

\paragraph{Episodic Memory.}
To capture context-dependent patterns that the transition table misses, we store full trajectory segments from past episodes in an episodic memory~$\mathcal{E}$~\citep{lewis2020rag,zheng2024synapse,kagaya2024rap,wang2024awm}. Each episode~$e$ is a trajectory segment from a completed task:

\begin{equation}
    e = (\textit{context}, \textit{actions}, \textit{observation},
         \textit{arg\_sources}, \textit{outcome}, \textit{lesson})
\end{equation}

where \emph{context} is a natural-language summary of the agent's current situation, \emph{actions} is the list of actions with full arguments, \emph{observation} is the environment response, \emph{arg\_sources} maps each argument value to its provenance by substring matching against prior tool observations, \emph{outcome} indicates task success, and \emph{lesson} is a takeaway auto-generated from episode metadata via template rules. All episode fields are constructed by rule-based heuristics without auxiliary LLM calls, meaning memory updates add zero inference latency. At speculation time, the current context is embedded using OpenAI's \texttt{text-embedding-3-small} model (1536 dimensions) and the top-$k$ most similar episodes are retrieved via cosine similarity, providing the speculator with structured examples of similar past situations. The only retrieval-time API call is this embedding query ($<$50\,ms), which runs during environment idle time alongside the speculator. We also record \emph{speculation-miss episodes}~\citep{ding2026agenther,ouyang2025reasoningbank} that store the predicted versus actual action, giving the speculator explicit examples of its own past errors. Finally, retrieval-time pruning inspired by Evo-Memory~\citep{wei2025evo} did not improve accuracy and is omitted from our reported results (see Appendix~\ref{app:pruning}).

\paragraph{Confusion Tracker.}
The confusion tracker identifies recurring prediction errors~\citep{shinn2023reflexion,zhao2024expel}. After each speculation verification, we record the (predicted\_action\_type, actual\_action\_type) pair. When a specific confusion pattern is observed $\ge 3$ times, it is surfaced as a hard constraint in the speculator's prompt, explicitly instructing the speculator to avoid the confused action type. Unlike episodic memory, which requires embedding retrieval, the confusion tracker is a lightweight dictionary of counts that provides immediate feedback as patterns cross the threshold.

All three memory systems are updated online after each completed task, so subsequent speculations benefit from accumulated experience. Speculation itself never affects the actor. The actor sees identical context regardless of prediction outcomes. Correct predictions use the cached result. Incorrect ones are discarded and execution proceeds normally, preserving the trajectory exactly as it would be without speculation.


\section{Experiments}
\label{sec:experiments}

\subsection{Setup}
\label{sec:setup}
We evaluate on six benchmarks spanning four agent modalities: customer service, web navigation, embodied tasks, and multi-hop question answering (\Cref{tab:benchmark_overview}). The benchmarks are grouped by speculation type: two benchmarks per type.

\begin{table}[t]
\centering
\caption{Benchmark characteristics grouped by speculation type.}
\label{tab:benchmark_overview}
\small
\begin{tabular}{lllccll}
\toprule
\textbf{Type}
  & \textbf{Benchmark}
  & \textbf{Metric}
  & \textbf{Tasks}
  & \textbf{Win}
  & \textbf{Action space}
  & \textbf{Pattern type} \\
\midrule
\multirow{2}{*}{\rotatebox[origin=c]{0}{\textbf{1}}}
  & WebArena
  & RO action match
  & 812
  & 52.3\%
  & Browser actions
  & Page navigation \\
  & VWA
  & RO action match
  & 910
  & 54.9\%
  & Browser actions
  & Visual navigation \\
\midrule
\multirow{2}{*}{\rotatebox[origin=c]{0}{\textbf{2}}}
  & ALFWorld
  & Obs.\ match
  & 134
  & 100\%
  & 13 verbs
  & Task-type templates \\
  & PDDL
  & Obs.\ match
  & 60
  & 83.3\%
  & Grid commands
  & State transitions \\
\midrule
\multirow{2}{*}{\rotatebox[origin=c]{0}{\textbf{3}}}
  & $\tau^{2}$-bench
  & RO action match
  & 164
  & 74.4\%
  & $\sim$30 APIs
  & API call sequences \\
  & HotpotQA
  & Action match
  & 300
  & 67.0\%
  & 3 ReAct actions
  & Multi-hop QA \\
\bottomrule
\end{tabular}
\end{table}

\paragraph{Benchmarks.}
\emph{Type~1 (action prediction):} WebArena~\citep{zhou2024webarena} tests web navigation across shopping, forum, and content management sites using 812~tasks. VisualWebArena (VWA)~\citep{koh2024vwa} extends WebArena with visual grounding on 910~tasks. \emph{Type~2 (observation prediction):} ALFWorld~\citep{shridhar2021alfworld} is a household task benchmark with a discrete 13-verb action space over 134~\texttt{valid\_unseen} episodes. PDDL planning tasks~\citep{vallati2015ipc,ma2024agentboard} require agents to navigate grid worlds and solve structured puzzles across 60~tasks. \emph{Type~3 (chained prediction):} $\tau^{2}$-bench~\citep{barres2025tau2} is a dual-control customer-service benchmark; we use both the retail and airline domains, totaling 164~tasks. HotpotQA~\citep{yang2018hotpotqa} is a multi-hop question answering benchmark where a ReAct agent~\citep{yao2023react} iteratively searches Wikipedia over 300~tasks.

\paragraph{Evaluation.}
Each speculation type determines the evaluation metric. For Type~1, we evaluate read-only action full match; for Type~2, observation exact match. For Type~3, the speculator predicts the environment response and then the agent's next action; we use action full match as the primary metric since it determines pre-launchability. We report \emph{aggregate accuracy}: total correct steps divided by total steps, pooled across all tasks. For $\tau^{2}$-bench, we exclude turn~0 from the denominator since the user's identity is initially unknown.

\paragraph{Models.}
Actor trajectories are collected once and replayed, eliminating actor variance across settings and isolating API costs to the speculator. The actor model is GPT-5.4 for ALFWorld, PDDL, $\tau^{2}$-bench, and HotpotQA, and GPT-5-mini for WebArena and VWA. The primary speculator is GPT-4.1-mini. We also leverage LLMs for visualization of experimental results.

\paragraph{Ablation settings.}
\label{sec:setup:settings} We evaluate seven progressive memory settings, all running in parallel during environment idle time with $n=3$ speculator instances per setting: (1)~\textbf{Stateless}: conversation history only, no memory; (2)~\textbf{Confusion}: adds confusion pattern constraints. To build cumulatively, all subsequent settings include this confusion tracking as a base layer: (3)~\textbf{Table}: adds transition table statistics; (4)~\textbf{Episodic}: replaces the table with retrieved past episodes to isolate retrieval effects; (5)~\textbf{Table+Episodic}: combines table and episodic memory; (6)~\textbf{Episodic+Miss}: identical to Episodic but adds speculation-miss episodes; (7)~\textbf{Full}: all components combined. A best-of-$k$ evaluation checks whether any instance predicted correctly. The prediction targets match the speculation types: Type~1 predicts actions, Type~2 predicts observations, and Type~3 chains both.

\subsection{Main Results}
\label{sec:main_results}

\begin{table*}[t]
\centering
\caption{Main results: aggregate accuracy (\%) and improvement over Stateless
($\Delta$, pp) across all six benchmarks ($k=1$).
Columns are grouped by speculation type.
Best value per benchmark in \textbf{bold}.
Cell shading indicates magnitude of $\Delta$.
All Stateless$\to$Best improvements are significant ($p < 0.001$,
McNemar's test over all evaluation steps; see
\Cref{tab:speculator_variance}).}
\label{tab:main_results}
\setlength{\tabcolsep}{3.5pt}
\begin{tabular}{lcccccccccccc}
\toprule
& \multicolumn{4}{c}{\textbf{Type~1: Action pred.}} &
  \multicolumn{4}{c}{\textbf{Type~2: Obs.\ pred.}} &
  \multicolumn{4}{c}{\textbf{Type~3: Chained}} \\
\cmidrule(lr){2-5} \cmidrule(lr){6-9} \cmidrule(lr){10-13}
& \multicolumn{2}{c}{WebArena} & \multicolumn{2}{c}{VWA}
& \multicolumn{2}{c}{ALFWorld} & \multicolumn{2}{c}{PDDL}
& \multicolumn{2}{c}{$\tau^{2}$-bench} & \multicolumn{2}{c}{HotpotQA} \\
\cmidrule(lr){2-3} \cmidrule(lr){4-5} \cmidrule(lr){6-7}
\cmidrule(lr){8-9} \cmidrule(lr){10-11} \cmidrule(lr){12-13}
& Acc & $\Delta$ & Acc & $\Delta$
& Acc & $\Delta$ & Acc & $\Delta$
& Acc & $\Delta$ & Acc & $\Delta$ \\
\midrule
Stateless
  & 19.8 & ---
  & 12.5 & ---
  & 16.3 & ---
  & 17.6 & ---
  & 12.7 & ---
  & 20.5 & --- \\
Confusion
  & 21.4 & \posL{+1.6}
  & 15.3 & \pos{+2.8}
  & 16.0 & \nega{$-$0.3}
  & 19.0 & \posL{+1.4}
  & 13.1 & \posL{+0.4}
  & 21.4 & \posL{+0.9} \\
Table
  & 21.9 & \pos{+2.1}
  & 15.8 & \pos{+3.3}
  & 19.5 & \pos{+3.2}
  & 21.1 & \pos{+3.5}
  & 15.2 & \pos{+2.5}
  & 21.1 & \posL{+0.6} \\
Episodic
  & 23.4 & \pos{+3.6}
  & \textbf{17.4} & \pos{\textbf{+4.9}}
  & 23.8 & \posM{+7.5}
  & 19.0 & \posL{+1.4}
  & 17.0 & \pos{+4.3}
  & 25.0 & \pos{+4.5} \\
Table+Episodic
  & 23.5 & \pos{+3.7}
  & 16.6 & \pos{+4.1}
  & 28.1 & \posM{+11.8}
  & 20.0 & \pos{+2.4}
  & 17.6 & \pos{+4.9}
  & \textbf{27.0} & \posM{\textbf{+6.5}} \\
Episodic+Miss
  & \textbf{23.7} & \pos{\textbf{+3.9}}
  & 16.7 & \pos{+4.2}
  & 38.6 & \posH{+22.3}
  & 33.0 & \posH{+15.4}
  & 16.8 & \pos{+4.1}
  & 26.0 & \posM{+5.5} \\
Full
  & 23.5 & \pos{+3.7}
  & 16.2 & \pos{+3.7}
  & \textbf{40.0} & \posH{\textbf{+23.7}}
  & \textbf{33.8} & \posH{\textbf{+16.2}}
  & \textbf{19.9} & \posM{\textbf{+7.2}}
  & \textbf{27.5} & \posM{\textbf{+7.0}} \\
\bottomrule
\end{tabular}
\end{table*}

\Cref{tab:main_results} summarizes speculation accuracy across all six benchmarks and seven ablation settings at $k$=1. Memory-augmented speculation improves over the Stateless baseline on every benchmark, with the best setting varying by domain and speculation type.

\paragraph{Type~1: Action prediction (WebArena, VWA).}
On WebArena, Episodic+Miss achieves 23.7\% read-only full match accuracy, +3.9\,pp absolute over Stateless at 19.8\%. The confusion tracker, transition table, and episodic memory each contribute, with the Stateless$\to$Confusion step at +1.6\,pp and Table$\to$Episodic swap at +1.5\,pp providing the largest incremental gains. Miss episodes add only +0.3\,pp, indicating that web navigation errors are less systematic than those on embodied tasks. On VWA, Episodic achieves 17.4\%, +4.9\,pp over Stateless. The confusion tracker alone provides +2.8\,pp, the largest single-step gain. Adding the transition table on top of episodic memory decreases accuracy by 0.8\,pp, reinforcing that episodic memory alone is preferable when action spaces are diverse.

\paragraph{Type~2: Observation prediction (ALFWorld, PDDL).}
On ALFWorld, Full achieves 40.0\%, $2.5\times$ the Stateless baseline at 16.3\%. This is the largest relative improvement across all benchmarks, reflecting ALFWorld's highly repetitive action sequences. Isolating each component's contribution, miss episodes provide the largest gain at +14.8\,pp when added to Episodic (Episodic$\to$Episodic+Miss, 23.8$\to$38.6\%), episodic memory adds +7.8\,pp over Confusion alone (Confusion$\to$Episodic, 16.0$\to$23.8\%), and the transition table adds +3.5\,pp (Confusion$\to$Table, 16.0$\to$19.5\%). On PDDL, Full achieves 33.8\%, $1.9\times$ Stateless. Failure learning dominates. Adding miss episodes to Episodic yields +14.0\,pp (Episodic$\to$Episodic+Miss, 19.0$\to$33.0\%), the largest single-component gain across all benchmarks, suggesting that planning tasks generate highly systematic prediction errors that miss episodes effectively correct.

\paragraph{Type~3: Chained prediction ($\tau^{2}$-bench, HotpotQA).}
On $\tau^{2}$-bench, Full achieves 19.9\%, +7.2\,pp over Stateless. Each memory component contributes positively: confusion adds +0.4\,pp, the transition table adds +2.1\,pp over confusion (Confusion$\to$Table, 13.1$\to$15.2\%), and adding episodic memory on top of the table yields a further +2.4\,pp (Table$\to$Table+Episodic, 15.2$\to$17.6\%). On HotpotQA, Full achieves 27.5\%, +7.0\,pp over Stateless. The gain is driven by action prediction, which requires structural reasoning about when to search, look up, or finish. These are exactly the workflow patterns that episodic memory captures.

Across benchmarks, each memory component has a distinct profile. The confusion tracker helps on five of six benchmarks, with the largest effect on VWA (+2.8\,pp), but slightly reduces accuracy on ALFWorld where prediction errors are less systematic. The transition table benefits embodied benchmarks with regular action sequences and also helps on $\tau^{2}$-bench (+2.1\,pp), though its gains are smaller than episodic memory on most benchmarks. Episodic memory provides consistent gains across all benchmarks by supplying argument values and observation templates from similar past situations. While the best memory settings achieve 17--40\% accuracy depending on the benchmark, which may appear low in isolation, speculative execution is lossless. Incorrect predictions are discarded at zero functional cost, meaning only the hit rate matters. At 20\% accuracy on a benchmark with $\ell_{\text{hit}} = 7$\,s, memory hides $\sim$140\,s of latency per 100 steps compared to non-speculative execution, and $\sim$50\,s more than the stateless baseline. The cost-benefit analysis in \Cref{sec:cost_benefit} quantifies these savings across all benchmarks. The speculator call is one to two orders of magnitude cheaper than the actor, so speculation yields a net benefit at any accuracy above $\sim$5\%.

\section{Analysis}
\label{sec:analysis}

\subsection{Experience Accumulation}

\begin{figure*}[t]
\centering

\begin{minipage}[t]{0.38\textwidth}
\centering{\scriptsize\textsc{Type 1: Action pred.}}
\end{minipage}
\hspace{-40pt}
\begin{minipage}[t]{0.38\textwidth}
\centering{\scriptsize\textsc{Type 2: Obs.\ pred.}}
\end{minipage}
\hspace{-30pt}
\begin{minipage}[t]{0.38\textwidth}
\centering{\scriptsize\textsc{Type 3: Chained pred.}}
\end{minipage}

\vspace{2pt}


\begin{minipage}[t]{0.325\textwidth}
\centering
\begin{tikzpicture}
\begin{axis}[
    width=1.12\textwidth,
    height=0.82\textwidth,
    ylabel={Cum.\ accuracy (\%)},
    ymin=12, ymax=35,
    legend style={at={(0.97,0.97)}, anchor=north east, font=\scriptsize,
                  draw=none, fill=white, fill opacity=0.85},
    ymajorgrids=true,
    grid style={gray!20, line width=0.3pt},
    title={\small\textbf{WebArena}},
    title style={yshift=-2pt},
    tick label style={font=\tiny},
    label style={font=\scriptsize},
    line width=1.2pt,
]
\addplot[color=cStateless, mark=none]
  coordinates {(50,23.6) (100,21.8) (150,20.0) (200,20.5) (250,18.0) (300,17.5) (350,17.1) (400,17.5) (450,17.5) (500,17.6) (550,18.6) (600,18.9) (650,19.6) (700,19.4) (750,19.6) (811,19.6)};
\addlegendentry{Stateless}
\addplot[color=cFull, mark=none]
  coordinates {(50,31.3) (100,30.9) (150,28.9) (200,30.3) (250,26.6) (300,26.2) (350,25.1) (400,25.4) (450,24.5) (500,23.9) (550,24.4) (600,24.6) (650,25.0) (700,24.5) (750,24.7) (811,24.6)};
\addlegendentry{Memory}
\end{axis}
\end{tikzpicture}
\end{minipage}
\hspace{-10pt}
\begin{minipage}[t]{0.325\textwidth}
\centering
\begin{tikzpicture}
\begin{axis}[
    width=1.12\textwidth,
    height=0.82\textwidth,
    ymin=8, ymax=45,
    ymajorgrids=true,
    grid style={gray!20, line width=0.3pt},
    title={\small\textbf{ALFWorld}},
    title style={yshift=-2pt},
    tick label style={font=\tiny},
    label style={font=\scriptsize},
    line width=1.2pt,
]
\addplot[color=cStateless, mark=none]
  coordinates {(1,16.7) (10,10.7) (20,11.2) (30,16.6) (40,18.6) (50,18.5) (60,19.0) (70,18.6) (80,18.4) (90,17.7) (100,17.5) (110,17.2) (120,16.9) (130,16.4) (134,16.3)};
\addplot[color=cFull, mark=none]
  coordinates {(1,16.7) (10,37.1) (20,42.4) (30,40.7) (40,41.5) (50,41.0) (60,40.9) (70,39.4) (80,38.6) (90,38.1) (100,39.0) (110,39.2) (120,39.8) (130,39.8) (134,40.0)};
\end{axis}
\end{tikzpicture}
\end{minipage}
\hspace{-10pt}
\begin{minipage}[t]{0.325\textwidth}
\centering
\begin{tikzpicture}
\begin{axis}[
    width=1.12\textwidth,
    height=0.82\textwidth,
    ymin=5, ymax=25,
    ymajorgrids=true,
    grid style={gray!20, line width=0.3pt},
    title={\small$\boldsymbol{\tau^{2}}$\textbf{-bench}},
    title style={yshift=-2pt},
    tick label style={font=\tiny},
    label style={font=\scriptsize},
    line width=1.2pt,
]
\addplot[color=cStateless, mark=none]
  coordinates {(10,17.1) (20,20.7) (30,10.3) (40,9.2) (50,7.5) (60,7.0) (70,6.8) (80,6.9) (90,8.5) (100,9.3) (110,9.5) (120,10.5) (130,11.2) (140,11.7) (150,12.0) (160,12.3) (164,12.7)};
\addplot[color=cFull, mark=none]
  coordinates {(10,22.9) (20,20.7) (30,11.2) (40,12.5) (50,11.8) (60,11.1) (70,12.1) (80,12.4) (90,13.9) (100,16.6) (110,16.4) (120,18.7) (130,18.6) (140,18.5) (150,18.1) (160,18.5) (164,19.9)};
\end{axis}
\end{tikzpicture}
\end{minipage}

\vspace{2pt}


\begin{minipage}[t]{0.325\textwidth}
\centering
\begin{tikzpicture}
\begin{axis}[
    width=1.12\textwidth,
    height=0.82\textwidth,
    xlabel={Task index},
    ylabel={Cum.\ accuracy (\%)},
    ymin=8, ymax=22,
    ymajorgrids=true,
    grid style={gray!20, line width=0.3pt},
    title={\small\textbf{VWA}},
    title style={yshift=-2pt},
    tick label style={font=\tiny},
    label style={font=\scriptsize},
    line width=1.2pt,
]
\addplot[color=cStateless, mark=none]
  coordinates {(50,12.0) (100,13.9) (150,14.3) (200,13.3) (250,12.3) (300,13.1) (350,13.9) (400,15.1) (450,15.1) (500,14.3) (550,13.9) (600,13.0) (650,12.9) (700,12.5) (750,12.6) (800,12.8) (850,13.0) (909,12.4)};
\addplot[color=cFull, mark=none]
  coordinates {(50,13.0) (100,13.9) (150,15.2) (200,15.0) (250,14.9) (300,15.6) (350,16.5) (400,17.6) (450,17.4) (500,16.8) (550,16.4) (600,16.2) (650,16.1) (700,15.7) (750,15.7) (800,16.1) (850,16.4) (909,16.0)};
\end{axis}
\end{tikzpicture}
\end{minipage}
\hspace{-10pt}
\begin{minipage}[t]{0.325\textwidth}
\centering
\begin{tikzpicture}
\begin{axis}[
    width=1.12\textwidth,
    height=0.82\textwidth,
    xlabel={Task index},
    ymin=10, ymax=40,
    ymajorgrids=true,
    grid style={gray!20, line width=0.3pt},
    title={\small\textbf{PDDL}},
    title style={yshift=-2pt},
    tick label style={font=\tiny},
    label style={font=\scriptsize},
    line width=1.2pt,
]
\addplot[color=cStateless, mark=none]
  coordinates {(5,22.5) (10,24.5) (15,21.1) (20,19.1) (25,18.6) (30,17.8) (35,16.5) (40,15.5) (45,16.5) (50,16.4) (55,16.8) (59,17.6)};
\addplot[color=cFull, mark=none]
  coordinates {(5,30.0) (10,29.1) (15,27.1) (20,25.6) (25,25.4) (30,24.4) (35,24.7) (40,27.0) (45,28.5) (50,28.3) (55,31.7) (59,33.8)};
\end{axis}
\end{tikzpicture}
\end{minipage}
\hspace{-10pt}
\begin{minipage}[t]{0.325\textwidth}
\centering
\begin{tikzpicture}
\begin{axis}[
    width=1.12\textwidth,
    height=0.82\textwidth,
    xlabel={Task index},
    ymin=15, ymax=32,
    ymajorgrids=true,
    grid style={gray!20, line width=0.3pt},
    title={\small\textbf{HotpotQA}},
    title style={yshift=-2pt},
    tick label style={font=\tiny},
    label style={font=\scriptsize},
    line width=1.2pt,
]
\addplot[color=cStateless, mark=none]
  coordinates {(20,23.6) (40,23.1) (60,22.7) (80,20.5) (100,18.6) (120,18.4) (140,19.1) (160,19.9) (180,20.2) (200,19.8) (220,19.7) (240,20.2) (260,20.0) (280,20.3) (300,20.5)};
\addplot[color=cFull, mark=none]
  coordinates {(20,27.3) (40,26.9) (60,26.2) (80,26.9) (100,24.8) (120,23.9) (140,24.4) (160,25.5) (180,26.5) (200,25.9) (220,27.3) (240,27.8) (260,27.1) (280,27.3) (300,27.5)};
\end{axis}
\end{tikzpicture}
\end{minipage}

\caption{Cumulative accuracy ($k=1$) across all six benchmarks, grouped by
speculation type. Each point shows aggregate accuracy over all tasks up to
that index. ``Memory'' denotes the best-performing memory setting per
benchmark. The learning effect is strongest on Type~2 benchmarks, where
memory climbs steadily while the stateless baseline remains flat.
All seven settings are shown in Appendix~\ref{app:all_learning_curves}.}
\label{fig:learning_curve}
\end{figure*}
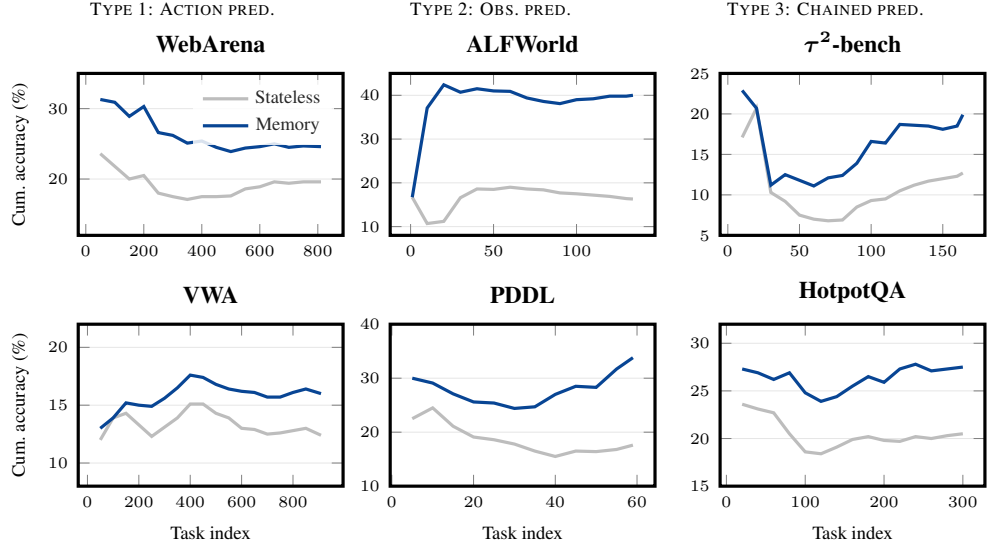

\Cref{fig:learning_curve} plots cumulative accuracy as tasks accumulate across all six benchmarks, comparing the stateless speculator against the best-performing memory setting per benchmark. The accumulation effect varies by speculation type. On Type~2 benchmarks, where repetitive action sequences provide reusable patterns, accuracy grows steadily with experience. On ALFWorld, memory quickly climbs to $\sim$40\% and stabilizes, while Stateless stays flat around 16\%. On PDDL, memory rises from $\sim$24\% to 34\% in the second half, while Stateless remains below 18\%. On Type~1 benchmarks, memory maintains a consistent 3--5\,pp advantage throughout without a strong upward trend, indicating that gains come primarily from the memory content itself rather than continued accumulation. On Type~3 benchmarks, memory maintains a consistent advantage on $\tau^{2}$-bench, reaching 19.9\% vs.\ 12.7\% for Stateless. HotpotQA shows a growing gap, with Full reaching 27.5\% vs.\ 20.5\% for Stateless, driven by action-pattern recognition rather than accumulating observation examples. Results are generally robust to task presentation order, with the Stateless baseline varying by at most 1.6\,pp and the top-ranked setting shifting only on benchmarks with smaller evaluation sets (see Appendix~\ref{app:ordering}).

\subsection{Speculator Model Sensitivity}
\begin{table}[t]
\centering
\caption{Speculator model comparison: aggregate accuracy (\%) of the best
memory setting and memory improvement ($\Delta$: Stateless$\to$best, pp)
across three speculator models. Cost is relative to GPT-4.1-mini.
WebArena and VWA use GPT-5-mini as actor; GPT-5.4-mini is omitted for
these benchmarks as it is comparable in capability to the actor.}
\label{tab:speculator_comparison}
\small
\resizebox{\columnwidth}{!}{
\begin{tabular}{lcccccccccccc}
\toprule
& \multicolumn{2}{c}{\textbf{WebArena}} &
  \multicolumn{2}{c}{\textbf{VWA}} &
  \multicolumn{2}{c}{\textbf{ALFWorld}} &
  \multicolumn{2}{c}{\textbf{PDDL}} &
  \multicolumn{2}{c}{\textbf{$\tau^{2}$-bench}} &
  \multicolumn{2}{c}{\textbf{HotpotQA}} \\
\cmidrule(lr){2-3} \cmidrule(lr){4-5} \cmidrule(lr){6-7}
\cmidrule(lr){8-9} \cmidrule(lr){10-11} \cmidrule(lr){12-13}
& Acc & $\Delta$ & Acc & $\Delta$ & Acc & $\Delta$
& Acc & $\Delta$ & Acc & $\Delta$ & Acc & $\Delta$ \\
\midrule
GPT-4.1-mini
  & 23.7 & \pos{+3.9} & 17.4 & \pos{+4.9}
  & 40.0 & \posH{+23.7} & 33.8 & \posH{+16.2}
  & 19.9 & \posM{+7.2} & 27.5 & \posM{+7.0} \\
GPT-5.4-mini
  & --- & --- & --- & ---
  & 51.3 & \posH{+18.9} & 51.5 & \posH{+27.2}
  & 24.2 & \pos{+4.2} & 32.8 & \posL{+2.1} \\
GPT-5.4-nano
  & 16.3 & \posM{+9.0} & 9.1 & \pos{+4.9}
  & 46.7 & \posH{+20.8} & 41.0 & \posH{+21.0}
  & 30.3 & \posM{+6.4} & 23.4 & \pos{+4.3} \\
\bottomrule
\end{tabular}
}
\end{table}
\Cref{tab:speculator_comparison} compares three speculator models of varying capability and cost. We omit GPT-5.4-mini for WebArena and VWA because its capability is comparable to their actor model, GPT-5-mini. The central finding is that memory improvement is model-agnostic. The Stateless-to-best delta is consistent across all tested models, ranging from +2--7\,pp on $\tau^{2}$-bench and HotpotQA to +16--27\,pp on ALFWorld and PDDL. Model capability primarily affects observation prediction, where GPT-5.4-mini improves over GPT-4.1-mini by +11\,pp on ALFWorld and +18\,pp on PDDL. The cheapest model, GPT-5.4-nano at $0.6\times$ the cost, achieves competitive memory deltas across all benchmarks, making it a highly cost-effective speculator.

\subsection{Latency Analysis}
\label{sec:cost_benefit}
\begin{figure}[t]
\centering

\begin{minipage}[t]{0.48\textwidth}
\centering
\begin{tikzpicture}
\begin{axis}[
    width=\textwidth,
    height=0.82\textwidth,
    xlabel={\small Environment delay (s)},
    ylabel={\small Time saved (s\,/\,100 steps)},
    ymin=0, ymax=250,
    xmin=0, xmax=22,
    xtick={0,5,10,15,20},
    ytick={0,50,100,150,200,250},
    legend style={at={(0.97,0.5)}, anchor=east, font=\scriptsize,
                  draw=none, fill=white, fill opacity=0.9,
                  row sep=-1pt},
    ymajorgrids=true,
    grid style={gray!20, line width=0.3pt},
    tick label style={font=\scriptsize},
    label style={font=\small},
    title={\small\textbf{(a) ALFWorld: savings vs.\ env delay}},
    title style={yshift=-2pt},
]
\addplot[color=cGray, line width=1.2pt, mark=none, densely dotted]
  coordinates {(0,0) (2,81.9) (3,116.6) (5,159.6) (7,184.5) (10,207.0) (12,215.8) (15,223.5) (20,231.0)};
\addlegendentry{Upper bound}

\addplot[color=cBlueDark, line width=1.2pt, mark=none, dashed]
  coordinates {(0,0) (2,0) (3,11.1) (5,83.2) (7,134.6) (10,172.0) (12,185.6) (15,196.7) (20,204.8)};
\addlegendentry{Predicted}

\addplot[color=cOrange, line width=1.5pt, mark=*, mark size=1.5pt]
  coordinates {(0,0) (2,2.2) (3,20.8) (5,86.7) (7,138.7) (10,182.6) (12,199.3) (15,214.2) (20,226.4)};
\addlegendentry{Measured}

\draw[<->, red, line width=0.6pt] (axis cs:3,20.8) -- (axis cs:3,116.6)
  node[midway, anchor=north west, xshift=8pt, font=\small, text=red] {spec overhead};

\end{axis}
\end{tikzpicture}
\end{minipage}%
\hfill
\begin{minipage}[t]{0.48\textwidth}
\centering
\begin{tikzpicture}
\begin{axis}[
    width=\textwidth,
    height=0.82\textwidth,
    xlabel={\small Step depth},
    ylabel={\small Median latency (s)},
    ymin=0, ymax=12,
    xmin=0.5, xmax=7.5,
    xtick={1,2,3,4,5,6,7},
    legend style={at={(0.03,0.97)}, anchor=north west, font=\tiny,
                  draw=none, fill=white, fill opacity=0.9,
                  row sep=-1pt},
    ymajorgrids=true,
    grid style={gray!20, line width=0.3pt},
    tick label style={font=\scriptsize},
    label style={font=\small},
    title={\small\textbf{(b) HotpotQA: latency and accuracy}},
    title style={yshift=-2pt},
    axis y line*=left,
]
\addplot[color=cBlueDark, line width=1.2pt, mark=*, mark size=1.2pt]
  coordinates {(1,4.0) (2,3.4) (3,2.8) (4,4.2) (5,7.2) (6,9.4) (7,7.2)};
\addlegendentry{Actor (GPT-5.4)}

\addplot[color=cOrange, line width=1.2pt, mark=square*, mark size=1.2pt]
  coordinates {(1,1.7) (2,1.6) (3,1.6) (4,1.6) (5,1.6) (6,1.8) (7,1.9)};
\addlegendentry{Spec (GPT-4.1-mini)}
\end{axis}

\begin{axis}[
    width=\textwidth,
    height=0.82\textwidth,
    ylabel={\small Action accuracy (\%)},
    ymin=0, ymax=70,
    xmin=0.5, xmax=7.5,
    xtick=\empty,
    ytick={0,20,40,60},
    legend style={at={(0.97,0.97)}, anchor=north east, font=\tiny,
                  draw=none, fill=white, fill opacity=0.9,
                  row sep=-1pt},
    tick label style={font=\scriptsize},
    label style={font=\small},
    axis y line*=right,
    axis x line=none,
]
\addplot[color=cGray, line width=1.0pt, mark=o, mark size=1.2pt, densely dotted]
  coordinates {(1,32.7) (2,19.2) (3,11.0) (4,9.5) (5,4.8) (6,10.5) (7,3.1)};
\addlegendentry{Stateless acc.}

\addplot[color=cBlueMed, line width=1.2pt, mark=triangle*, mark size=1.5pt, dashed]
  coordinates {(1,33.0) (2,33.6) (3,22.0) (4,14.3) (5,7.1) (6,18.4) (7,9.4)};
\addlegendentry{Memory acc.}
\end{axis}
\end{tikzpicture}
\end{minipage}

\caption{\textbf{Live latency validation.} \textbf{(a)}~ALFWorld savings vs.\ delay $\ell_{\mathrm{env}}$. Dotted: theoretical bound (zero-latency speculator). At $\ell_{\mathrm{env}} < 3$\,s, speculator latency ($\sim$2.7\,s) constrains gains (red). Measured results (solid) track replay predictions (dashed), validating the offline model. \textbf{(b)}~HotpotQA latency and accuracy by step. Speculator is $2\text{--}5\times$ faster than actor, providing a consistent head start. Memory accuracy (dashed) consistently outperforms Stateless (dotted) across all depths.}

\label{fig:live_latency}
\end{figure}
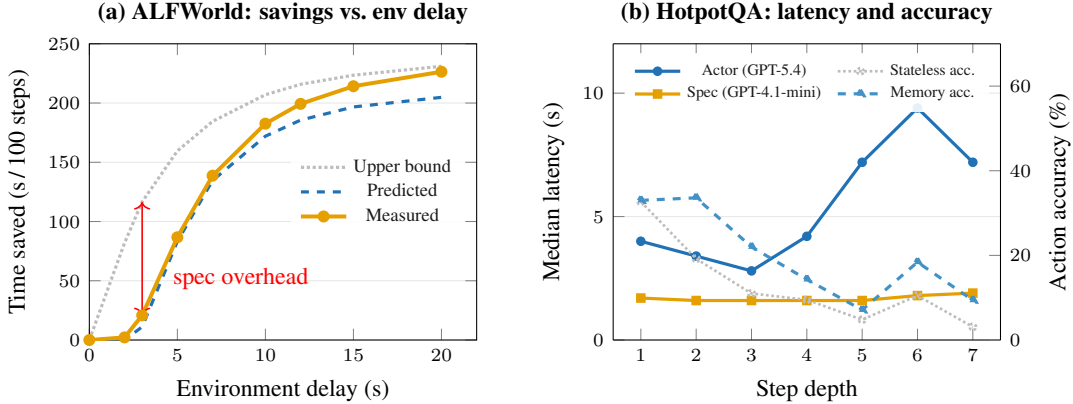
\paragraph{Cost and theoretical savings.}
The speculator LLM call is one to two orders of magnitude cheaper than the actor across all benchmarks.\footnote{Pricing:
GPT-4.1-mini \$0.40/\$1.60 per MTok (input/output);
GPT-5.4 \$2.50/\$15.00 per MTok.
Source: \texttt{openai.com/api/pricing}, May 2026.} For Types~2 and~3, each speculation additionally pre-launches an actor call, but on a hit the pre-launched result substitutes for the next step's actor call, so the net expected additional cost is $k \cdot C_{\text{S}} + (1 - \text{acc}) \cdot C_{\text{act}}$. \Cref{tab:cost_benefit} in Appendix~\ref{app:cost} reports the full cost-latency breakdown per 100 steps for all six benchmarks.

\paragraph{Live validation.}
To validate that replay-measured accuracy translates to real-world savings, we run the actor and speculator live on the 300 tasks of HotpotQA and 134 episodes of ALFWorld, recording per-step latencies alongside prediction accuracy. On HotpotQA, the GPT-5.4 actor has an overall median latency of 4.9\,s, whereas the GPT-4.1-mini speculator finishes in 1.7\,s, yielding a $\sim$3\,s head start per hit. The live action hit rate is 40.0\% across 610 speculated steps, yielding a $1.05\times$ analytical speedup and saving 892\,s over a 19,024\,s baseline. On ALFWorld, the actor has an overall median latency of 4.5\,s and the speculator 2.7\,s, with a 44.9\% observation hit rate across 1,638 speculated steps. Because ALFWorld evaluates observation prediction, time savings depend on the environment feedback latency $\ell_{\text{env}}$. As \Cref{fig:live_latency}(a) shows, the measured savings closely track the replay-based prediction across environment delays from 2\,s to 20\,s, confirming that offline accuracy reliably predicts online savings. However, three factors reduce these practical savings below the theoretical upper bound of an instant speculator. First, the 2.7\,s median latency of the speculator itself consumes a portion of the delay window. At small delays where $\ell_{\text{env}} < 3$\,s, most of this window is spent executing the speculator rather than building a head start, as highlighted by the red annotation in \Cref{fig:live_latency}(a). Second, variance in API latency causes some individual speculator calls to exceed the environment delay even if the median time remains well within it. Third, successful predictions on the final step of a task cannot save time because there is no subsequent action to pre-launch. Together, these effects explain the gap between the theoretical upper bound and the measured performance, a gap that steadily diminishes as $\ell_{\text{env}}$ increases.

\subsection{Step-Depth Accuracy}
\label{sec:step_depth}
Memory-augmented speculation maintains its accuracy advantage across step depths on all six benchmarks, as detailed in \Cref{fig:step_depth} of Appendix~\ref{app:step_depth}. The advantage is largest at early steps where workflow patterns are most predictable, but persists at later steps where episodic retrieval supplies relevant templates. \Cref{fig:live_latency}(b) connects accuracy to latency: the speculator remains near 1.7\,s regardless of depth, whereas the actor varies from roughly 3 to 9.5\,s. Consequently, each correct speculation at deeper steps saves more wall-clock time. As plotted on the right axis, the full memory setting maintains a consistent accuracy advantage over the stateless baseline at every depth. The combined effect of higher accuracy and a larger latency gap makes deeper steps particularly valuable targets for speculation.


\section{Related Work}
\label{sec:related}

\paragraph{Speculative and parallel execution for agents.}
\label{sec:related:spec}
Tool-using LLM agents~\citep{yao2023react,schick2023toolformer,wang2024opendevin} execute actions sequentially, making latency proportional to the number of tool calls. \citet{ye2025speculative} propose Speculative Actions, predicting the agent's next tool call during idle time with a stateless speculator. Other approaches target different bottlenecks: LLMCompiler~\citep{kim2024llm} parallelizes independent calls within a single turn via a task-dependency DAG, which is complementary to our cross-turn speculation. SpecReason~\citep{pan2025specreason} and ConfSpec~\citep{liu2026confspec} apply speculative decoding at the internal reasoning-step level, whereas our work operates at the environment-action level. Most closely related is PASTE~\citep{sui2026paste}, which learns DAG-structured tool-calling patterns to guide speculation. Unlike PASTE, our approach retrieves contextually similar past trajectories with concrete argument values, learns from prediction failures online, and updates all memory dynamically during deployment.

\paragraph{Memory systems for agents.}
\label{sec:related:memory}
A growing body of work equips agents with non-parametric memory updated from past trajectories. ExpeL~\citep{zhao2024expel} extracts insights by contrasting successful and failed trajectory pairs, while AWM~\citep{wang2024awm} distills reusable action workflows from past episodes for retrieval at test time. Other systems store procedural abstractions~\citep{fang2025memp,wang2023voyager}, full trajectories as exemplars~\citep{zheng2024synapse,kagaya2024rap,sarukkai2025self}, or reasoning strategies derived from success and failure~\citep{ouyang2025reasoningbank}. Retrieval-augmented episodic memory~\citep{lewis2020rag,park2023generative,majumder2023clin,zhang2026memrl} and failure-aware learning~\citep{shinn2023reflexion,ding2026agenther,allard2026erl} have similarly become foundational to agent design. Recent work also addresses memory curation: EvolveR~\citep{wu2025evolver} applies explicit distill, deduplicate, update, and filter operations, whereas Memory-R1~\citep{yan2025memory} and AgeMem~\citep{yu2026agentic} train RL controllers over memory management actions. Crucially, all these systems target the \emph{actor}'s task performance; we apply this same family of techniques to improve the \emph{speculator}'s prediction accuracy.


\section{Conclusion}
\label{sec:conclusion}
We presented memory-augmented speculative execution, which equips LLM agent speculators with three complementary memory systems to learn from past trajectories. Evaluation across six diverse benchmarks shows that structured memory consistently improves accuracy for all speculation types. The largest gains occur in observation prediction, where repetitive environment responses provide highly reusable patterns. These model-agnostic improvements vary by domain: Type~2 tasks benefit from continuous experience accumulation, while Type~1 tasks rely primarily on retrieving specific past content. Because speculation runs during environment idle time, these accuracy gains translate directly to measured latency reductions without adding wall-clock cost to the actor. However, combining all memory components is not always optimal, and unpredictable elements like page-specific UI indices still bottleneck absolute accuracy. Future work includes confidence-based memory routing, structure-aware web prediction, and broader production deployments.

\begin{ack}
We thank the anonymous reviewers for their constructive feedback.
\end{ack}


\bibliographystyle{plainnat}
\bibliography{references}


\newpage
\appendix

\section{Additional Results}
\label{app:additional}

\subsection{Best-of-$k$ Scaling}
\label{app:bestofk}

\begin{figure*}[t]
\centering


\begin{minipage}[t]{0.325\textwidth}
\centering
\begin{tikzpicture}
\begin{axis}[
    width=1.12\textwidth,
    height=0.82\textwidth,
    ylabel={Accuracy (\%)},
    xtick={1,2,3},
    xmin=0.7, xmax=3.3,
    ymin=15, ymax=30,
    legend style={at={(0.03,0.97)}, anchor=north west, font=\scriptsize,
                  draw=none, fill=white, fill opacity=0.85},
    ymajorgrids=true,
    grid style={gray!20, line width=0.3pt},
    title={\small\textbf{WebArena}},
    title style={yshift=-2pt},
    tick label style={font=\tiny},
    label style={font=\scriptsize},
    line width=1.2pt,
]
\addplot[color=cStateless, mark=o, mark size=1.5pt, dashed]
  coordinates {(1,19.8) (2,21.8) (3,22.8)};
\addlegendentry{Stateless}
\addplot[color=cFull, mark=diamond*, mark size=1.5pt]
  coordinates {(1,23.5) (2,25.5) (3,26.5)};
\addlegendentry{Memory}
\end{axis}
\end{tikzpicture}
\end{minipage}
\hspace{-10pt}
\begin{minipage}[t]{0.325\textwidth}
\centering
\begin{tikzpicture}
\begin{axis}[
    width=1.12\textwidth,
    height=0.82\textwidth,
    xtick={1,2,3},
    xmin=0.7, xmax=3.3,
    ymin=12, ymax=48,
    ymajorgrids=true,
    grid style={gray!20, line width=0.3pt},
    title={\small\textbf{ALFWorld}},
    title style={yshift=-2pt},
    tick label style={font=\tiny},
    label style={font=\scriptsize},
    line width=1.2pt,
]
\addplot[color=cStateless, mark=o, mark size=1.5pt, dashed]
  coordinates {(1,16.3) (2,17.4) (3,17.7)};
\addplot[color=cFull, mark=diamond*, mark size=1.5pt]
  coordinates {(1,40.0) (2,42.6) (3,43.8)};
\end{axis}
\end{tikzpicture}
\end{minipage}
\hspace{-10pt}
\begin{minipage}[t]{0.325\textwidth}
\centering
\begin{tikzpicture}
\begin{axis}[
    width=1.12\textwidth,
    height=0.82\textwidth,
    xtick={1,2,3},
    xmin=0.7, xmax=3.3,
    ymin=8, ymax=28,
    ymajorgrids=true,
    grid style={gray!20, line width=0.3pt},
    title={\small$\boldsymbol{\tau^{2}}$\textbf{-bench}},
    title style={yshift=-2pt},
    tick label style={font=\tiny},
    label style={font=\scriptsize},
    line width=1.2pt,
]
\addplot[color=cStateless, mark=o, mark size=1.5pt, dashed]
  coordinates {(1,12.7) (2,15.2) (3,17.2)};
\addplot[color=cFull, mark=diamond*, mark size=1.5pt]
  coordinates {(1,19.9) (2,22.7) (3,24.4)};
\end{axis}
\end{tikzpicture}
\end{minipage}

\vspace{2pt}


\begin{minipage}[t]{0.325\textwidth}
\centering
\begin{tikzpicture}
\begin{axis}[
    width=1.12\textwidth,
    height=0.82\textwidth,
    xlabel={Speculators ($k$)},
    ylabel={Accuracy (\%)},
    xtick={1,2,3},
    xmin=0.7, xmax=3.3,
    ymin=8, ymax=22,
    ymajorgrids=true,
    grid style={gray!20, line width=0.3pt},
    title={\small\textbf{VWA}},
    title style={yshift=-2pt},
    tick label style={font=\tiny},
    label style={font=\scriptsize},
    line width=1.2pt,
]
\addplot[color=cStateless, mark=o, mark size=1.5pt, dashed]
  coordinates {(1,12.5) (2,14.3) (3,15.4)};
\addplot[color=cFull, mark=diamond*, mark size=1.5pt]
  coordinates {(1,16.2) (2,18.2) (3,19.4)};
\end{axis}
\end{tikzpicture}
\end{minipage}
\hspace{-10pt}
\begin{minipage}[t]{0.325\textwidth}
\centering
\begin{tikzpicture}
\begin{axis}[
    width=1.12\textwidth,
    height=0.82\textwidth,
    xlabel={Speculators ($k$)},
    xtick={1,2,3},
    xmin=0.7, xmax=3.3,
    ymin=12, ymax=42,
    ymajorgrids=true,
    grid style={gray!20, line width=0.3pt},
    title={\small\textbf{PDDL}},
    title style={yshift=-2pt},
    tick label style={font=\tiny},
    label style={font=\scriptsize},
    line width=1.2pt,
]
\addplot[color=cStateless, mark=o, mark size=1.5pt, dashed]
  coordinates {(1,17.6) (2,19.0) (3,20.6)};
\addplot[color=cFull, mark=diamond*, mark size=1.5pt]
  coordinates {(1,33.8) (2,37.7) (3,39.1)};
\end{axis}
\end{tikzpicture}
\end{minipage}
\hspace{-10pt}
\begin{minipage}[t]{0.325\textwidth}
\centering
\begin{tikzpicture}
\begin{axis}[
    width=1.12\textwidth,
    height=0.82\textwidth,
    xlabel={Speculators ($k$)},
    xtick={1,2,3},
    xmin=0.7, xmax=3.3,
    ymin=16, ymax=36,
    ymajorgrids=true,
    grid style={gray!20, line width=0.3pt},
    title={\small\textbf{HotpotQA}},
    title style={yshift=-2pt},
    tick label style={font=\tiny},
    label style={font=\scriptsize},
    line width=1.2pt,
]
\addplot[color=cStateless, mark=o, mark size=1.5pt, dashed]
  coordinates {(1,20.5) (2,24.2) (3,25.8)};
\addplot[color=cFull, mark=diamond*, mark size=1.5pt]
  coordinates {(1,27.5) (2,30.9) (3,33.1)};
\end{axis}
\end{tikzpicture}
\end{minipage}

\caption{Best-of-$k$ scaling from $k=1$ to $k=3$ across all six benchmarks. Running $k=3$ parallel speculators improves accuracy by 1.4--5.6\,pp over $k=1$ across all domains. Memory-informed settings benefit significantly more from scaling than the Stateless baseline on observation-prediction benchmarks (e.g., ALFWorld Full +3.8\,pp vs.\ Stateless +1.4\,pp), suggesting that memory-guided predictions explore a more diverse set of plausible environment continuations.}
\label{fig:bestofk_scaling}
\end{figure*}
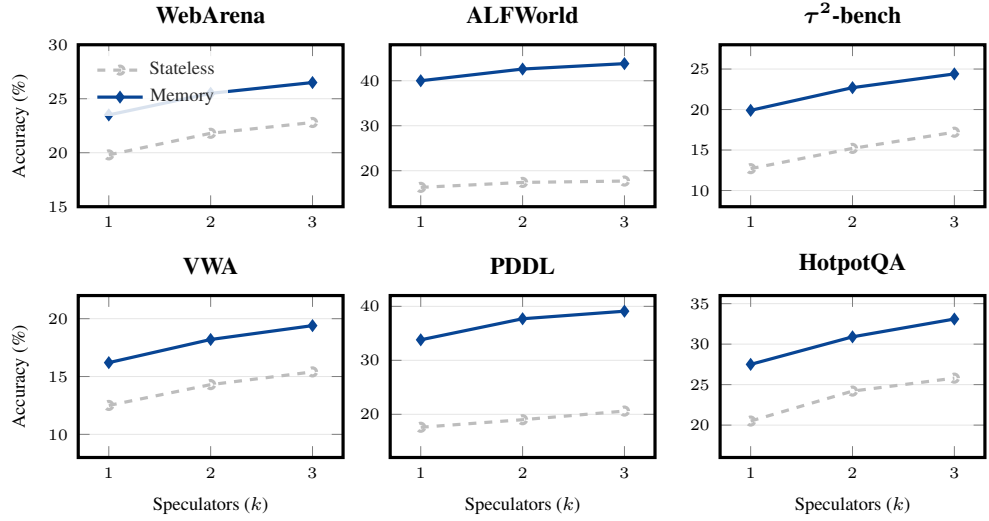

\Cref{fig:bestofk_scaling} illustrates the accuracy scaling from $k=1$ to $k=3$ speculators across all six benchmarks. Two primary patterns emerge. First, increasing the number of speculators yields consistent gains across the board, ranging from a +1.4\,pp increase for ALFWorld Stateless to +5.6\,pp for HotpotQA Memory. Second, memory-informed settings generally benefit more from scaling than the stateless baseline on observation-prediction benchmarks. For instance, on ALFWorld, the Full setting gains +3.8\,pp compared to +1.4\,pp for Stateless. This divergence suggests that memory-guided predictions explore a more diverse set of plausible environment continuations rather than producing near-identical guesses.

\subsection{Won vs.\ Lost Trajectory Split}
\label{app:wonlost}

\begin{table}[h]
\centering
\caption{Won vs.\ lost trajectory split: aggregate accuracy (\%, $k=1$)
for Stateless and best Memory settings. Won trajectories generally achieve 
higher speculation accuracy, with the gap largest on HotpotQA (+14.0\,pp 
for Stateless, +16.1\,pp for Memory). ALFWorld is omitted as all episodes 
succeeded.}
\label{tab:won_lost}
\small
\resizebox{\columnwidth}{!}{
\begin{tabular}{lccccccc}
\toprule
& & & & \multicolumn{2}{c}{\textbf{Won}} & \multicolumn{2}{c}{\textbf{Lost}} \\
\cmidrule(lr){5-6} \cmidrule(lr){7-8}
Benchmark & \#Won & \#Lost & Total & Stateless & Memory & Stateless & Memory \\
\midrule
WebArena & 425 & 387 & 812 & 24.9 & 28.3 & 16.9 & 21.0 \\
VWA      & 500 & 410 & 910 & 14.8 & 19.7 & 11.1 & 16.1 \\
PDDL     & 50  & 10  & 60  & 20.4 & 33.3 & 10.5 & 35.0 \\
$\tau^{2}$-bench & 122 & 42 & 164 & 14.9 & 21.2 & 7.8  & 17.0 \\
HotpotQA & 201 & 99  & 300 & 26.7 & 34.6 & 12.7 & 18.5 \\
\bottomrule
\end{tabular}
}
\end{table}

\Cref{tab:won_lost} analyzes speculation accuracy based on task outcome across five benchmarks; ALFWorld is omitted as all episodes were successful. Successful trajectories generally yield higher speculation accuracy, suggesting that they follow more predictable, standard patterns. This gap is most pronounced on HotpotQA, where successful tasks achieve 26.7\% Stateless accuracy compared to 12.7\% for failures, a $2.1\times$ difference. This suggests that failed tasks involve more exploratory or erratic search behaviors that are inherently harder to speculate.

A similar trend holds for WebArena, VWA, and $\tau^{2}$-bench. PDDL is a slight outlier, showing marginally higher Memory accuracy on lost tasks (35.0\% vs.\ 33.3\% for won). However, the limited number of failures in PDDL ($n=10$) makes this estimate less reliable than the other benchmarks. Overall, the data confirms that speculation is most effective when the agent is on a successful, predictable path.

\subsection{Ordering Effects}
\label{app:ordering}

\begin{table*}[t]
\centering
\caption{Ordering effects on aggregate accuracy (\%, $k=1$) across
three task orderings: \emph{sequential} (default), \emph{shuffled}
(random permutation), and \emph{grouped} (clustered by type). The
rightmost column shows the Stateless baseline's maximum variation.
The top-ranked memory setting shifts on some benchmarks, but the
absolute differences between top-performing settings remain small.}
\label{tab:ordering_effects}
\small
\setlength{\tabcolsep}{2.5pt}
\resizebox{\textwidth}{!}{
\begin{tabular}{llcccccccr}
\toprule
Benchmark & Ordering & Stateless & Confusion & Table & Episodic & Tbl+Epi & Epi+Miss & Full & Range \\
\midrule
\multirow{3}{*}{WebArena}
  & Sequential & 19.8 & \posL{21.4} & \pos{21.9} & \pos{23.4} & \pos{23.5} & \pos{\textbf{23.7}} & \pos{23.5} & \\
  & Shuffled   & 19.8 & \posL{21.5} & \pos{22.1} & \pos{\textbf{24.2}} & \pos{23.9} & \pos{23.6} & \pos{23.7} & \\
  & Grouped    & 19.6 & \posL{21.5} & \pos{22.3} & \pos{23.5} & \pos{\textbf{23.8}} & \pos{23.6} & \pos{23.8} & 0.2 \\
\midrule
\multirow{3}{*}{VWA}
  & Sequential & 12.5 & \pos{15.3} & \pos{15.8} & \pos{\textbf{17.4}} & \pos{16.6} & \pos{16.7} & \pos{16.2} & \\
  & Shuffled   & 12.7 & \posL{14.7} & \pos{16.1} & \pos{\textbf{17.6}} & \pos{17.1} & \pos{16.9} & \pos{16.8} & \\
  & Grouped    & 12.7 & \posL{14.9} & \pos{15.4} & \pos{\textbf{17.1}} & \pos{16.2} & \pos{16.9} & \pos{16.4} & 0.2 \\
\midrule
\multirow{3}{*}{ALFWorld}
  & Sequential & 16.3 & \nega{16.0} & \pos{19.5} & \posM{23.8} & \posM{28.1} & \posH{38.6} & \posH{\textbf{40.0}} & \\
  & Shuffled   & 16.0 & {16.0} & \pos{19.9} & \posM{26.2} & \posM{29.2} & \posH{39.6} & \posH{\textbf{42.4}} & \\
  & Grouped    & 15.5 & \posL{16.3} & \pos{19.9} & \posM{22.7} & \posM{26.5} & \posH{34.8} & \posH{\textbf{38.0}} & 0.8 \\
\midrule
\multirow{3}{*}{PDDL}
  & Sequential & 17.6 & \posL{19.0} & \pos{21.1} & \posL{19.0} & \pos{20.0} & \posH{33.0} & \posH{\textbf{33.8}} & \\
  & Shuffled   & 16.1 & \posL{17.9} & \pos{20.6} & \pos{19.4} & \pos{20.8} & \posH{33.9} & \posH{\textbf{36.9}} & \\
  & Grouped    & 16.0 & \pos{19.3} & \pos{20.4} & \pos{19.9} & \pos{20.2} & \posH{33.5} & \posH{\textbf{34.9}} & 1.6 \\
\midrule
\multirow{3}{*}{$\tau^{2}$-bench}
  & Sequential & 12.7 & \posL{13.1} & \pos{15.2} & \pos{17.0} & \pos{17.6} & \pos{16.8} & \posM{\textbf{19.9}} & \\
  & Shuffled   & 11.9 & \pos{14.3} & \pos{15.6} & \pos{16.0} & \posM{17.2} & \posM{17.2} & \posM{\textbf{17.6}} & \\
  & Grouped    & 12.3 & \posL{14.3} & \posL{14.3} & \posM{\textbf{18.0}} & \pos{16.0} & \pos{16.6} & \posM{17.4} & 0.8 \\
\midrule
\multirow{3}{*}{HotpotQA}
  & Sequential & 20.5 & \posL{21.4} & \posL{21.1} & \pos{25.0} & \posM{27.0} & \posM{26.0} & \posM{\textbf{27.5}} & \\
  & Shuffled   & 20.0 & \posL{21.1} & \posL{22.0} & \posM{27.8} & \posM{26.9} & \posM{27.4} & \posM{\textbf{28.8}} & \\
  & Grouped    & 19.8 & \posL{21.5} & \posL{21.4} & \posM{27.0} & \posM{\textbf{27.5}} & \posM{25.6} & \posM{25.8} & 0.7 \\
\bottomrule
\end{tabular}
}
\end{table*}

\Cref{tab:ordering_effects} reports aggregate accuracy across three task orderings on all six benchmarks. Three primary findings emerge from this analysis.

\paragraph{Stability of the Stateless baseline.}
The Stateless baseline is remarkably stable, varying by at most 1.6\,pp across orderings on PDDL and by less than 1\,pp on five of the six benchmarks. Since the Stateless setting uses no memory, this minor variation reflects only the inherent difference in predictability among specific task sequences within the benchmarks.

\paragraph{Bounded variation in memory settings.}
Observation-prediction benchmarks (ALFWorld and PDDL) show the highest sensitivity to task ordering when memory is enabled. For instance, ALFWorld Full ranges from 38.0\% in the grouped ordering to 42.4\% when shuffled, representing a 4.4\,pp spread. The grouped ordering, which clusters similar task types, likely produces less diverse episodic memory during the early stages of a run. Conversely, the shuffled ordering performs best on these benchmarks, suggesting that early exposure to a diverse set of examples improves the quality of subsequent episodic retrieval.

\paragraph{Rank shifts among top settings.}
The optimal memory configuration is consistent for some benchmarks but shifts for others. On VWA, the Episodic setting is consistently best across all orderings, while the Full setting remains top-ranked for ALFWorld and PDDL. On WebArena, the top-ranked setting alternates between Episodic, Tbl+Epi, and Epi+Miss, though the absolute difference between these winners is less than 0.6\,pp. On $\tau^{2}$-bench and HotpotQA, the best setting shifts more substantially, with specific configurations varying by up to 3.0\,pp (e.g., HotpotQA Full). Despite these rank shifts, the Stateless-to-best gap remains positive across all orderings for every benchmark, confirming that memory-augmented speculation is robust to task presentation order.

\subsection{Cost--Benefit Analysis}
\label{app:cost}

\begin{table*}[t]
\caption{Cost-benefit analysis, all values per 100 agent steps. $\ell_{\text{hit}}$ is wall-clock saving per correct speculation; $T_{100}$ is baseline latency without speculation. $C_{\text{act}}$, $C_{\text{S}}$, and $C_{\text{M}}$ represent actor and speculator costs. Spec and +Mem columns denote latency reduction in seconds. Cell shading intensity is proportional to the absolute reduction. Speculation occurs at every step except the first of each task.}
\label{tab:cost_benefit}
\centering
\small
\setlength{\tabcolsep}{3.5pt}
\resizebox{\textwidth}{!}{
\begin{tabular}{lc r r ccc rr rr rr}
\toprule
& & & & & & & \multicolumn{2}{c}{$k=1$}
                         & \multicolumn{2}{c}{$k=2$}
                         & \multicolumn{2}{c}{$k=3$} \\
\cmidrule(lr){8-9} \cmidrule(lr){10-11} \cmidrule(lr){12-13}
Benchmark & Type
  & $\ell_{\text{hit}}$
  & $T_{100}$
  & $C_{\text{act}}$
  & $C_{\text{S}}$
  & $C_{\text{M}}$
  & Spec & {+Mem}
  & Spec & {+Mem}
  & Spec & {+Mem} \\
\midrule
WebArena    & 1 & 12.3\,s & 3600\,s
  & 0.35 & 0.032 & 0.044
  & $-$244\,s & \cellcolor{cPosH!18}$-$292\,s
  & $-$288\,s & \cellcolor{cPosH!20}$-$333\,s
  & $-$311\,s & \cellcolor{cPosH!22}$-$355\,s \\
VWA         & 1 & 17.1\,s & 4197\,s
  & 0.44 & 0.032 & 0.044
  & $-$213\,s & \cellcolor{cPosH!16}$-$297\,s
  & $-$244\,s & \cellcolor{cPosH!18}$-$325\,s
  & $-$264\,s & \cellcolor{cPosH!20}$-$343\,s \\
\midrule
ALFWorld    & 2 & 7.1\,s & 710\,s
  & 0.21 & 0.012 & 0.024
  & $-$116\,s & \cellcolor{cPosH!42}$-$284\,s
  & $-$124\,s & \cellcolor{cPosH!44}$-$303\,s
  & $-$126\,s & \cellcolor{cPosH!46}$-$311\,s \\
PDDL        & 2 & 6.3\,s & 630\,s
  & 0.23 & 0.022 & 0.034
  & $-$111\,s & \cellcolor{cPosH!30}$-$213\,s
  & $-$120\,s & \cellcolor{cPosH!33}$-$238\,s
  & $-$130\,s & \cellcolor{cPosH!35}$-$246\,s \\
\midrule
$\tau^{2}$-bench  & 3 & 12.5\,s & 4000\,s
  & 1.65 & 0.025 & 0.037
  & $-$159\,s & \cellcolor{cPosH!30}$-$248\,s
  & $-$190\,s & \cellcolor{cPosH!35}$-$284\,s
  & $-$215\,s & \cellcolor{cPosH!38}$-$305\,s \\
HotpotQA    & 3 & 10.0\,s & 1000\,s
  & 0.28 & 0.023 & 0.035
  & $-$205\,s & \cellcolor{cPosH!32}$-$275\,s
  & $-$242\,s & \cellcolor{cPosH!37}$-$309\,s
  & $-$258\,s & \cellcolor{cPosH!40}$-$331\,s \\
\bottomrule
\end{tabular}
}
\end{table*}

\Cref{tab:cost_benefit} provides a per-benchmark cost-benefit breakdown. All values are normalized to 100 agent steps for direct comparison. The columns are constructed as follows. The wall-clock time saved per correct speculation, $\ell_{\text{hit}}$, is determined by the speculation type as described in \Cref{sec:method:framework}. For Type~1, $\ell_{\text{hit}} = \ell_{\text{env}}$ represents the environment processing time saved by pre-launching the predicted action. These values are measured from trajectory timestamps, yielding 12.3\,s for WebArena and 17.1\,s for VWA, which includes visual processing overhead. For Type~2, $\ell_{\text{hit}} = \min(\ell_{\text{LLM}},\; \ell_{\text{env}} - \ell_{\text{spec}})$. The table uses representative deployment latencies where $\ell_{\text{env}} \approx 7$\,s for embodied tasks. For Type~3, $\ell_{\text{hit}} = \ell_{\text{LLM}} - \ell_{\text{spec}}$, representing the head start gained by the speculator finishing before the actor. $T_{100}$ is the baseline total latency for 100 steps without speculation, computed from the actor latencies recorded during trajectory collection. $C_{\text{act}}$ is the actor LLM cost for 100 steps, computed from token counts in the collected trajectories and the provider's pricing (see footnote in \Cref{sec:cost_benefit}). $C_{\text{S}}$ and $C_{\text{M}}$ are the speculator costs without and with memory context. Memory adds 200--500 extra input tokens per call, increasing cost by \$0.01--0.02 per 100 steps. The Spec and +Mem columns show the estimated latency reduction calculated as $\text{acc} \times \ell_{\text{hit}} \times 100$, where acc is the aggregate accuracy from \Cref{tab:main_results}.

Representative API latencies from our provider include $\ell_{\text{LLM}} \approx 5\text{--}10$\,s for GPT-5.4, $\ell_{\text{spec}} \approx 1\text{--}3$\,s for GPT-4.1-mini, and $\ell_{\text{env}}$ ranging from less than 0.1\,s for text simulators to approximately 17\,s for VWA. The speculator call is $10\text{--}70\times$ cheaper than the actor call, so speculation yields a net benefit at any accuracy above roughly 5\%. Memory systems improve accuracy at negligible incremental cost because embedding retrieval adds less than 50\,ms per turn.

\subsection{All-Setting Learning Curves}
\label{app:all_learning_curves}

\input{figures/cumulative_accuracy}

\Cref{fig:learning_curve} in the main text compares the stateless baseline against the best-performing memory configuration for each benchmark. \Cref{fig:cumulative_accuracy} plots all seven configurations to show the full performance progression as experience accumulates across the task sequence. On ALFWorld and PDDL, the configurations separate into two clear tiers. The Epi+Miss and Memory settings form an upper tier that diverges sharply from the remaining five settings after roughly 40 tasks. This separation demonstrates that learning from prediction failures is the dominant performance factor for observation-prediction benchmarks. On WebArena and VWA, all memory settings cluster more tightly. In these domains, Episodic retrieval provides the largest initial lift, while additional components offer diminishing returns. On $\tau^{2}$-bench, the Memory setting gradually pulls ahead of all other configurations during the second half of the trajectory. On HotpotQA, Tbl+Epi and Memory track closely throughout the run, and both outperform the other configurations by approximately 7\,pp.

\subsection{Per-Group Breakdown}
\label{app:per_task}

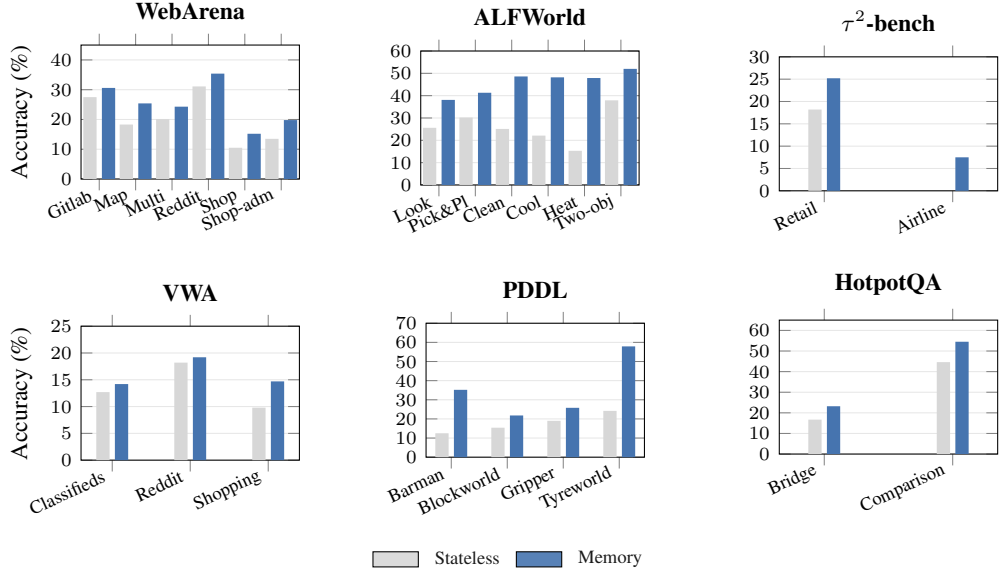
\begin{figure*}[t]
\centering

\pgfplotsset{
  group bar base/.style={
    ybar,
    ymajorgrids=true,
    grid style={gray!20, line width=0.3pt},
    tick label style={font=\scriptsize},
    label style={font=\small},
    area legend,
    ymin=0,
    legend style={font=\scriptsize, draw=none, fill=white,
                  fill opacity=0.9, column sep=4pt},
  },
}


\begin{minipage}[t]{0.32\textwidth}\centering
\begin{tikzpicture}
\begin{axis}[
    group bar base,
    width=\textwidth,
    height=0.75\textwidth,
    bar width=5pt,
    ylabel={Accuracy (\%)},
    symbolic x coords={Gitlab,Map,Multi,Reddit,Shop,Shop-adm},
    xtick=data,
    xticklabel style={font=\scriptsize, rotate=25, anchor=east},
    ymax=45,
    ytick={0,10,20,30,40},
    enlarge x limits=0.10,
    title={\small\textbf{WebArena}},
    legend to name=sharedlegend,
    legend columns=2,
]
\addplot[fill=cStateless!60, draw=none]
  coordinates {
    (Gitlab,27.5) (Map,18.3) (Multi,20.1)
    (Reddit,31.1) (Shop,10.5) (Shop-adm,13.5)
  };
\addlegendentry{Stateless}

\addplot[fill=cFull!75, draw=none]
  coordinates {
    (Gitlab,30.6) (Map,25.4) (Multi,24.3)
    (Reddit,35.4) (Shop,15.2) (Shop-adm,19.8)
  };
\addlegendentry{Memory}
\end{axis}
\end{tikzpicture}
\end{minipage}\hfill
%
\begin{minipage}[t]{0.32\textwidth}\centering
\begin{tikzpicture}
\begin{axis}[
    group bar base,
    width=\textwidth,
    height=0.75\textwidth,
    bar width=5pt,
    symbolic x coords={Look,Pick\&Pl,Clean,Cool,Heat,Two-obj},
    xtick=data,
    xticklabel style={font=\scriptsize, rotate=25, anchor=east},
    ymax=60,
    ytick={0,10,20,30,40,50,60},
    enlarge x limits=0.10,
    title={\small\textbf{ALFWorld}},
]
\addplot[fill=cStateless!60, draw=none]
  coordinates {
    (Look,25.6) (Pick\&Pl,30.2) (Clean,25.1)
    (Cool,22.1) (Heat,15.3) (Two-obj,37.9)
  };

\addplot[fill=cFull!75, draw=none]
  coordinates {
    (Look,38.1) (Pick\&Pl,41.3) (Clean,48.6)
    (Cool,48.2) (Heat,47.9) (Two-obj,52.0)
  };
\end{axis}
\end{tikzpicture}
\end{minipage}\hfill
%
\begin{minipage}[t]{0.32\textwidth}\centering
\begin{tikzpicture}
\begin{axis}[
    group bar base,
    width=\textwidth,
    height=0.75\textwidth,
    bar width=5pt,
    symbolic x coords={Retail,Airline},
    xtick=data,
    xticklabel style={font=\scriptsize, rotate=25, anchor=east},
    ymax=30,
    ytick={0,5,10,15,20,25,30},
    enlarge x limits=0.35,
    title={\small\textbf{$\tau^{2}$-bench}},
]
\addplot[fill=cStateless!60, draw=none]
  coordinates {(Retail,18.2) (Airline,0.0)};

\addplot[fill=cFull!75, draw=none]
  coordinates {(Retail,25.2) (Airline,7.5)};
\end{axis}
\end{tikzpicture}
\end{minipage}

\vspace{6pt}


\begin{minipage}[t]{0.32\textwidth}\centering
\begin{tikzpicture}
\begin{axis}[
    group bar base,
    width=\textwidth,
    height=0.75\textwidth,
    bar width=5pt,
    ylabel={Accuracy (\%)},
    symbolic x coords={Classifieds,Reddit,Shopping},
    xtick=data,
    xticklabel style={font=\scriptsize, rotate=25, anchor=east},
    ymax=25,
    ytick={0,5,10,15,20,25},
    enlarge x limits=0.20,
    title={\small\textbf{VWA}},
]
\addplot[fill=cStateless!60, draw=none]
  coordinates {(Classifieds,12.7) (Reddit,18.2) (Shopping,9.8)};

\addplot[fill=cFull!75, draw=none]
  coordinates {(Classifieds,14.2) (Reddit,19.2) (Shopping,14.7)};
\end{axis}
\end{tikzpicture}
\end{minipage}\hfill
%
\begin{minipage}[t]{0.32\textwidth}\centering
\begin{tikzpicture}
\begin{axis}[
    group bar base,
    width=\textwidth,
    height=0.75\textwidth,
    bar width=5pt,
    symbolic x coords={Barman,Blockworld,Gripper,Tyreworld},
    xtick=data,
    xticklabel style={font=\scriptsize, rotate=25, anchor=east},
    ymax=70,
    ytick={0,10,20,30,40,50,60,70},
    enlarge x limits=0.15,
    title={\small\textbf{PDDL}},
]
\addplot[fill=cStateless!60, draw=none]
  coordinates {
    (Barman,12.5) (Blockworld,15.4) (Gripper,19.0) (Tyreworld,24.2)
  };

\addplot[fill=cFull!75, draw=none]
  coordinates {
    (Barman,35.2) (Blockworld,21.8) (Gripper,25.8) (Tyreworld,57.9)
  };
\end{axis}
\end{tikzpicture}
\end{minipage}\hfill
%
\begin{minipage}[t]{0.32\textwidth}\centering
\begin{tikzpicture}
\begin{axis}[
    group bar base,
    width=\textwidth,
    height=0.75\textwidth,
    bar width=5pt,
    symbolic x coords={Bridge,Comparison},
    xtick=data,
    xticklabel style={font=\scriptsize, rotate=25, anchor=east},
    ymax=65,
    ytick={0,10,20,30,40,50,60},
    enlarge x limits=0.35,
    title={\small\textbf{HotpotQA}},
]
\addplot[fill=cStateless!60, draw=none]
  coordinates {(Bridge,16.7) (Comparison,44.6)};

\addplot[fill=cFull!75, draw=none]
  coordinates {(Bridge,23.2) (Comparison,54.5)};
\end{axis}
\end{tikzpicture}
\end{minipage}

\vspace{4pt}

\ref{sharedlegend}

\caption{Per-group accuracy breakdown ($k$=1) across all six benchmarks.
Groups correspond to websites for WebArena/VWA, task types for ALFWorld,
domains for $\tau^{2}$-bench, planning domains for PDDL, and question
types for HotpotQA. Memory improves over Stateless across nearly all
groups, with the largest gains on ALFWorld task types with stereotyped
sequences and PDDL Tyreworld.}
\label{fig:per_group_breakdown}
\end{figure*}

\Cref{fig:per_group_breakdown} details the per-group accuracy for the Stateless and Memory settings across all six benchmarks. On ALFWorld, the Memory setting improves accuracy across all six task types. The largest gains occur on Heat tasks with a +32.6\,pp increase and Cool tasks with a +26.1\,pp increase. These tasks feature repetitive action sequences that are highly amenable to learning from past trajectories. Similarly, on PDDL, Tyreworld benefits dramatically with a +33.7\,pp gain, whereas Blockworld sees a more modest improvement of +6.4\,pp because of its more variable state transitions. WebArena shows consistent improvement across all six websites, led by Reddit with a +4.3\,pp absolute gain. VWA also shows moderate, uniform gains across its three site categories. On $\tau^{2}$-bench, the retail domain benefits substantially, achieving 25.2\% accuracy with Memory compared to 18.2\% for Stateless. The airline domain shows lower absolute accuracy due to its smaller training set and more diverse API patterns. On HotpotQA, comparison questions remain easier to predict than bridge questions for both settings, but Memory provides consistent gains for both types. These results on VWA and HotpotQA demonstrate that episodic retrieval provides stable benefits even across heterogeneous task groups.

\subsection{Step-Depth Accuracy}
\label{app:step_depth}

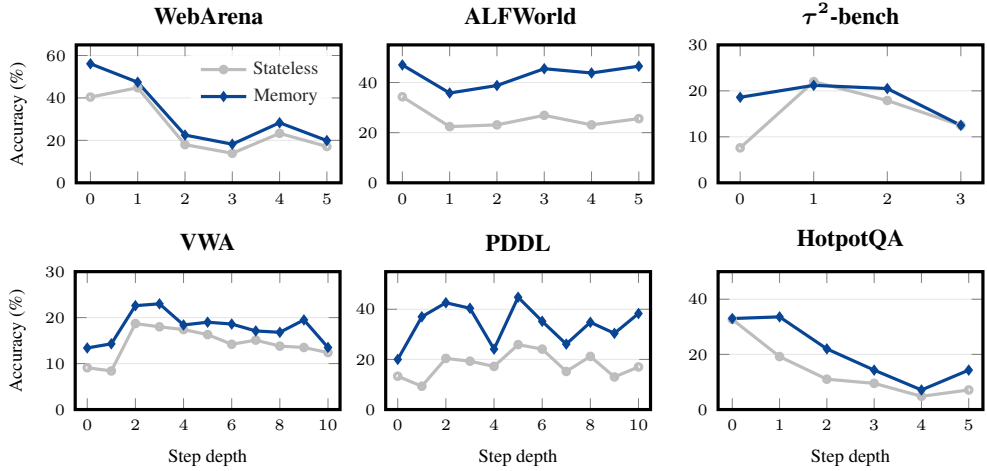
\begin{figure*}[t]
\centering


\begin{minipage}[t]{0.325\textwidth}
\centering
\begin{tikzpicture}
\begin{axis}[
    width=1.12\textwidth,
    height=0.75\textwidth,
    ylabel={Accuracy (\%)},
    ymin=0, ymax=65,
    xmin=-0.3, xmax=5.3,
    xtick={0,1,2,3,4,5},
    legend style={at={(0.97,0.97)}, anchor=north east, font=\scriptsize,
                  draw=none, fill=white, fill opacity=0.85},
    ymajorgrids=true,
    grid style={gray!20, line width=0.3pt},
    title={\small\textbf{WebArena}},
    title style={yshift=-2pt},
    tick label style={font=\tiny},
    label style={font=\scriptsize},
    line width=1.2pt,
]
\addplot[color=cStateless, mark=o, mark size=1.2pt]
  coordinates {(0,40.4) (1,44.8) (2,18.0) (3,13.9) (4,23.3) (5,17.1)};
\addlegendentry{Stateless}
\addplot[color=cFull, mark=diamond*, mark size=1.2pt]
  coordinates {(0,56.1) (1,47.4) (2,22.5) (3,18.2) (4,28.3) (5,19.9)};
\addlegendentry{Memory}
\end{axis}
\end{tikzpicture}
\end{minipage}
\hspace{-10pt}
\begin{minipage}[t]{0.325\textwidth}
\centering
\begin{tikzpicture}
\begin{axis}[
    width=1.12\textwidth,
    height=0.75\textwidth,
    ymin=0, ymax=55,
    xmin=-0.3, xmax=5.3,
    xtick={0,1,2,3,4,5},
    ymajorgrids=true,
    grid style={gray!20, line width=0.3pt},
    title={\small\textbf{ALFWorld}},
    title style={yshift=-2pt},
    tick label style={font=\tiny},
    label style={font=\scriptsize},
    line width=1.2pt,
]
\addplot[color=cStateless, mark=o, mark size=1.2pt]
  coordinates {(0,34.3) (1,22.4) (2,23.1) (3,26.9) (4,23.1) (5,25.6)};
\addplot[color=cFull, mark=diamond*, mark size=1.2pt]
  coordinates {(0,47.0) (1,35.8) (2,38.8) (3,45.5) (4,43.8) (5,46.5)};
\end{axis}
\end{tikzpicture}
\end{minipage}
\hspace{-10pt}
\begin{minipage}[t]{0.325\textwidth}
\centering
\begin{tikzpicture}
\begin{axis}[
    width=1.12\textwidth,
    height=0.75\textwidth,
    ymin=0, ymax=30,
    xmin=-0.3, xmax=3.3,
    xtick={0,1,2,3},
    ymajorgrids=true,
    grid style={gray!20, line width=0.3pt},
    title={\small$\boldsymbol{\tau^{2}}$\textbf{-bench}},
    title style={yshift=-2pt},
    tick label style={font=\tiny},
    label style={font=\scriptsize},
    line width=1.2pt,
]
\addplot[color=cStateless, mark=o, mark size=1.2pt]
  coordinates {(0,7.6) (1,22.0) (2,17.9) (3,12.5)};
\addplot[color=cFull, mark=diamond*, mark size=1.2pt]
  coordinates {(0,18.6) (1,21.2) (2,20.5) (3,12.5)};
\end{axis}
\end{tikzpicture}
\end{minipage}

\vspace{2pt}


\begin{minipage}[t]{0.325\textwidth}
\centering
\begin{tikzpicture}
\begin{axis}[
    width=1.12\textwidth,
    height=0.75\textwidth,
    xlabel={Step depth},
    ylabel={Accuracy (\%)},
    ymin=0, ymax=30,
    xmin=-0.5, xmax=10.5,
    xtick={0,2,4,6,8,10},
    ymajorgrids=true,
    grid style={gray!20, line width=0.3pt},
    title={\small\textbf{VWA}},
    title style={yshift=-2pt},
    tick label style={font=\tiny},
    label style={font=\scriptsize},
    line width=1.2pt,
]
\addplot[color=cStateless, mark=o, mark size=1.2pt]
  coordinates {(0,9.1) (1,8.4) (2,18.7) (3,18.0) (4,17.4) (5,16.3) (6,14.2) (7,15.1) (8,13.8) (9,13.5) (10,12.4)};
\addplot[color=cFull, mark=diamond*, mark size=1.2pt]
  coordinates {(0,13.4) (1,14.3) (2,22.6) (3,23.0) (4,18.4) (5,19.0) (6,18.6) (7,17.1) (8,16.8) (9,19.5) (10,13.5)};
\end{axis}
\end{tikzpicture}
\end{minipage}
\hspace{-10pt}
\begin{minipage}[t]{0.325\textwidth}
\centering
\begin{tikzpicture}
\begin{axis}[
    width=1.12\textwidth,
    height=0.75\textwidth,
    xlabel={Step depth},
    ymin=0, ymax=55,
    xmin=-0.5, xmax=10.5,
    xtick={0,2,4,6,8,10},
    ymajorgrids=true,
    grid style={gray!20, line width=0.3pt},
    title={\small\textbf{PDDL}},
    title style={yshift=-2pt},
    tick label style={font=\tiny},
    label style={font=\scriptsize},
    line width=1.2pt,
]
\addplot[color=cStateless, mark=o, mark size=1.2pt]
  coordinates {(0,13.3) (1,9.3) (2,20.4) (3,19.3) (4,17.2) (5,25.9) (6,24.1) (7,15.2) (8,21.2) (9,13.0) (10,17.0)};
\addplot[color=cFull, mark=diamond*, mark size=1.2pt]
  coordinates {(0,20.0) (1,37.0) (2,42.6) (3,40.4) (4,24.1) (5,44.8) (6,35.2) (7,26.1) (8,34.8) (9,30.4) (10,38.3)};
\end{axis}
\end{tikzpicture}
\end{minipage}
\hspace{-10pt}
\begin{minipage}[t]{0.325\textwidth}
\centering
\begin{tikzpicture}
\begin{axis}[
    width=1.12\textwidth,
    height=0.75\textwidth,
    xlabel={Step depth},
    ymin=0, ymax=50,
    xmin=-0.3, xmax=5.3,
    xtick={0,1,2,3,4,5},
    ymajorgrids=true,
    grid style={gray!20, line width=0.3pt},
    title={\small\textbf{HotpotQA}},
    title style={yshift=-2pt},
    tick label style={font=\tiny},
    label style={font=\scriptsize},
    line width=1.2pt,
]
\addplot[color=cStateless, mark=o, mark size=1.2pt]
  coordinates {(0,32.7) (1,19.2) (2,11.0) (3,9.5) (4,4.8) (5,7.1)};
\addplot[color=cFull, mark=diamond*, mark size=1.2pt]
  coordinates {(0,33.0) (1,33.6) (2,22.0) (3,14.3) (4,7.1) (5,14.3)};
\end{axis}
\end{tikzpicture}
\end{minipage}

\caption{Speculation accuracy by step depth across all six benchmarks.
Step depth~0 is the first speculated step (not the initial task step,
which is excluded).
``Memory'' denotes the best-performing memory setting per benchmark.
Memory maintains higher accuracy at most depths, with the advantage
largest on Type~2 benchmarks (15--20pp on ALFWorld, 10--20pp on PDDL).}
\label{fig:step_depth}
\end{figure*}

\Cref{fig:step_depth} illustrates how speculation accuracy evolves with step depth across all six benchmarks. On WebArena, accuracy is highest at step~0, where Memory reaches 56\% compared to 40\% for Stateless. Performance drops at step~2 as predictions shift from general navigational commands to more granular, element-specific interactions. ALFWorld maintains a consistent 15--20\,pp advantage across all depths, indicating that the benefit of memory is not restricted to initial steps. On PDDL, the Memory setting achieves 37--43\% accuracy at steps~1--3 while the Stateless baseline remains below 20\%, a gap that persists through step~10. VWA displays a steady 3--5\,pp Memory advantage across most depths. On $\tau^{2}$-bench, Stateless shows low initial accuracy, whereas Memory provides an 11\,pp advantage at step~0 and peaks at 24.5\% by step~2. Finally, HotpotQA exhibits a tighter separation, with Memory maintaining a 6--8\,pp advantage during intermediate reasoning steps.

\subsection{Episodic Memory Pruning}
\label{app:pruning}

We explored three retrieval-time pruning passes inspired by Evo-Memory~\citep{wei2025evo}. The first is a low-utility penalty that halves the similarity score for episodes from failed tasks where speculation was also incorrect. The second is redundancy deduplication, which removes candidates with an inter-episode cosine similarity above 0.92. The third is a miss-pattern penalty that reduces scores by $0.7\times$ for episodes whose primary action is mispredicted in at least three instances. The pruning process over-fetches by a factor of $3\times$ before filtering the results to $k$ candidates.

Across all benchmarks, pruning did not meaningfully change accuracy. On ALFWorld, the Memory setting increased from 40.0\% to 40.4\% with pruning enabled, while other benchmarks showed similarly negligible effects. We attribute this to two factors. First, the episodic memory is relatively small, containing hundreds rather than thousands of episodes, so redundancy is not yet a significant bottleneck. Second, the miss-pattern penalty can inadvertently suppress useful episodes if a frequently mispredicted action type also appears in successful retrieved records. Consequently, all results reported in this paper use the configuration with pruning disabled.

\subsection{Environment Delay Sensitivity (ALFWorld)}
\label{app:env_delay}

For Type~2 benchmarks, the speculator executes during the environment feedback window. While ALFWorld utilizes a text simulator with near-zero latency, target deployments such as household robotics involve physical execution delays. We measured actual actor and speculator latencies across all 134 episodes and analytically computed the speedup for a range of representative environment delays.

\begin{center}
\small
\setlength{\tabcolsep}{8pt}
\begin{tabular}{r c r r r c}
\toprule
$\ell_{\mathrm{env}}$ & Spec ready & Hits & Save/hit & Total saved & Speedup \\
\midrule
2\,s  & 16.9\% & 146 & 0.3\,s & 39\,s    & 1.003$\times$ \\
5\,s  & 90.5\% & 686 & 2.2\,s & 1,541\,s & 1.085$\times$ \\
7\,s  & 97.3\% & 723 & 3.4\,s & 2,465\,s & 1.119$\times$ \\
10\,s & 99.3\% & 732 & 4.4\,s & 3,244\,s & 1.128$\times$ \\
15\,s & 99.9\% & 735 & 5.2\,s & 3,806\,s & 1.113$\times$ \\
20\,s & 99.9\% & 735 & 5.5\,s & 4,023\,s & 1.095$\times$ \\
\bottomrule
\end{tabular}
\end{center}

As shown in the table, absolute time saved grows monotonically with the environment delay, increasing from 39\,s at a 2\,s delay to 4,023\,s at a 20\,s delay. However, the speedup ratio peaks near $\ell_{\mathrm{env}} = 10$\,s. This occurs because larger delays inflate the baseline latency faster than the savings grow. The per-hit savings, defined as $\min(\ell_{\mathrm{env}} - \ell_{\mathrm{spec}}, \ell_{\mathrm{actor}})$, are increasingly capped by the actor latency of the subsequent step rather than the environment delay window. In practical deployments, the environment latency is typically fixed by the hardware, making the speedup at that specific delay the relevant performance metric.

\section{Limitations and Broader Impact}
\label{app:limitations}
\subsection{Element Index Prediction}
\label{app:index_prediction}

The primary source of prediction error in web navigation benchmarks is element index prediction for \texttt{click\_element} actions. Element indices are arbitrary, page-specific identifiers assigned by the accessibility tree parser. These indices frequently change across different pages, page loads, or even within a single session as dynamic content renders. Because the speculator lacks access to the target page's DOM at the time of prediction, it cannot resolve which specific index corresponds to the intended element. This volatility fundamentally limits full-match accuracy on WebArena and VWA, where \texttt{click\_element} accounts for the majority of read-only actions. Incorporating lightweight page-structure features, such as a compressed DOM skeleton or element-role embeddings, into the speculation context represents a promising direction to mitigate this bottleneck.

\subsection{Memory Combination and Interference}
\label{app:interference}
The full memory configuration, which combines all three systems, underperforms simpler settings on benchmarks such as WebArena and VWA. For instance, adding the transition table to episodic memory on VWA reduces accuracy by 0.8\,pp. This suggests that aggregate transition statistics can sometimes mislead the speculator when action spaces are highly diverse. More broadly, naively concatenating all memory signals into the speculator's context can introduce redundancy, as episodic examples may already encode the relevant transition patterns, or conflicting guidance, such as when the transition table favors a frequent action that a miss episode explicitly warns against.

These observations motivate selective memory routing: a lightweight classifier or confidence-based mechanism that chooses the optimal memory configuration for each task or turn rather than always utilizing the full stack. Our evaluation of seven distinct settings provides the necessary training signal for such a router, as accuracy data for each configuration is available across various task types.

\subsection{Speculator Variance}
\label{app:variance}

\begin{table}[h]
\centering
\caption{Accuracy (\%) for the Stateless baseline and best memory configuration with mean $\pm$ standard deviation. For Type~1 and Type~2, columns $s_0, s_1, s_2$ represent three parallel speculators from the same run. For Type~3, columns represent three independent task orderings. The rightmost column shows $p$-values from McNemar's test for the Stateless-to-Memory improvement.}
\label{tab:speculator_variance}
\small
\setlength{\tabcolsep}{3.5pt}
\begin{tabular}{llccccc}
\toprule
Benchmark & Setting & $s_0$ & $s_1$ & $s_2$ & Mean $\pm$ Std & $p$ \\
\midrule
\multirow{2}{*}{WebArena}
  & Stateless      & 19.8 & 19.8 & 19.9 & 19.8 $\pm$ 0.05 & \\
  & Epi+Miss        & 23.7 & 23.9 & 23.9 & 23.8 $\pm$ 0.09 & $< 0.001$ \\
\midrule
\multirow{2}{*}{VWA}
  & Stateless      & 12.5 & 12.5 & 12.8 & 12.6 $\pm$ 0.12 & \\
  & Episodic        & 17.4 & 17.1 & 17.3 & 17.3 $\pm$ 0.10 & $< 0.001$ \\
\midrule
\multirow{2}{*}{ALFWorld}
  & Stateless      & 16.3 & 16.2 & 16.2 & 16.2 $\pm$ 0.05 & \\
  & Memory          & 40.0 & 41.5 & 41.4 & 41.0 $\pm$ 0.65 & $< 0.001$ \\
\midrule
\multirow{2}{*}{PDDL}
  & Stateless      & 17.6 & 16.0 & 17.1 & 16.9 $\pm$ 0.66 & \\
  & Memory          & 33.8 & 34.1 & 33.4 & 33.8 $\pm$ 0.28 & $< 0.001$ \\
\midrule
\multirow{2}{*}{$\tau^{2}$-bench}
  & Stateless      & 12.7 & 11.9 & 12.3 & 12.3 $\pm$ 0.33 & \\
  & Memory          & 19.9 & 17.6 & 17.4 & 18.3 $\pm$ 1.14 & $< 0.001$ \\
\midrule
\multirow{2}{*}{HotpotQA}
  & Stateless      & 20.5 & 20.0 & 19.8 & 20.1 $\pm$ 0.29 & \\
  & Memory          & 27.5 & 28.8 & 25.8 & 27.4 $\pm$ 1.23 & $< 0.001$ \\
\bottomrule
\end{tabular}
\end{table}

Each experiment utilizes three parallel speculators with identical prompts and temperatures. \Cref{tab:speculator_variance} reports the accuracy for the Stateless baseline and the best-performing memory configuration for each benchmark. In Type~1 and Type~2 benchmarks, three parallel speculators from the same execution show standard deviations ranging from 0.04\,pp to 0.66\,pp. For Type~3 benchmarks, three independent task orderings serve as replications to produce standard deviations between 0.29\,pp and 1.23\,pp. McNemar's tests conducted over all evaluation steps confirm that the accuracy improvement from the Stateless baseline to the Memory setting is statistically significant across all six benchmarks with $p < 0.001$. This test assesses whether the proportion of steps where memory succeeds but Stateless fails exceeds the reverse, providing a robust per-step significance assessment. Our replay-based evaluation protocol eliminates actor variance across settings because all configurations are evaluated against the identical actor trajectories. The $s_0$ values in \Cref{tab:speculator_variance} correspond exactly to the primary results reported in \Cref{tab:main_results}.

\subsection{Replay vs.\ Online Evaluation}
\label{app:replay_online}
Our replay-based design evaluates speculation accuracy in isolation. The speculator predicts against pre-recorded actor trajectories, ensuring that correct predictions do not actually accelerate the interaction or change the actor's context during the evaluation. In a live deployment, however, correct speculations would reduce latency and potentially allow the actor to process more turns within a fixed time budget. This live feedback loop could be positive if faster interactions lead to more coherent trajectories, but it could also be negative if a changed temporal context invalidates the replay assumption. A live deployment study incorporating real API latencies is necessary to measure true end-to-end wall-clock savings and validate that the accuracy measured during replay translates to a proportional speedup in practice.

\subsection{Broader Impact}
\label{app:impact}
Speculative execution for LLM agents is a latency-optimization technique that does not alter the agent's final behavior because the actor's decisions remain identical regardless of whether speculation is employed. The lossless guarantee ensures that incorrect predictions are silently discarded, introducing no new failure modes beyond those already present in the base agent. However, faster agent execution could amplify existing risks if the underlying agent is deployed in sensitive domains such as autonomous customer service or financial transactions without adequate safeguards. Because our memory systems store action-level trajectory data that could contain user information, production deployments should apply appropriate data retention policies and anonymization to the episodic memory store to protect user privacy.

\section{Prompt Templates}
\label{app:prompts}

\paragraph{Episodic memory format.}
Retrieved episodes are formatted as structured tuples injected into the
speculator's context.  The stored \emph{observation} field (Eq.~4) is used
for embedding-based retrieval but not surfaced in the prompt, as the
speculator already has the current conversation context.  The
\emph{arg\_sources} field is folded into the action formatting:
\begin{quote}
\small\ttfamily
--- Example 1 (similarity=0.87) ---\\
Situation: User asked to return items from order \#W6247578.\\
Agent actions:\\
~~get\_order\_details(order\_id=\#W6247578)\\
~~~~order\_id: from user message\\
Outcome: SUCCEEDED\\
Takeaway: Standard order lookup pattern after user provides order ID.
\end{quote}

\paragraph{Speculation-miss episode format.}
Miss episodes store the predicted versus actual action:
\begin{quote}
\small\ttfamily
--- Speculation miss (similarity: 0.82) ---\\
Pattern: Speculator predicted click(7) but agent actually used click(12).\\
Context: Product listing page with multiple items.\\
---
\end{quote}

\paragraph{Contrastive transition table format.}
The transition table is serialized as a ranked list of next-action
candidates with frequency and success rate, conditioned on the current
action:
\begin{quote}
\small\ttfamily
Historical patterns from 85 past tasks (71 success, 14 failure):\\
After get\_order\_details, the most likely next tools are:\\
~~1. get\_user\_details --- 42\% of the time (success rate: 88\%)\\
~~~~~typical args: order\_id,user\_id(63\%)\\
~~2. cancel\_pending\_order --- 25\% of the time (success rate: 72\%)\\
~~~~~typical args: order\_id,reason(91\%)\\
\\
Transitions to AVOID (high failure rate):\\
~~- modify\_pending\_order --- 12\% of the time but only 33\% success rate
\end{quote}
For benchmarks using turn-level speculation (ALFWorld), \texttt{get\_full\_context()}
provides the complete table rather than conditioning on a single action.

\paragraph{Confusion tracker format.}
Confusion patterns exceeding the threshold are injected as hard constraints:
\begin{quote}
\small\ttfamily
KNOWN PREDICTION ERRORS (avoid these):\\
- You predicted scroll\_down 5 times when the agent actually used\\
\phantom{-~}click\_element. Do NOT predict scroll\_down in this context.
\end{quote}



\end{document}